%% file: main.tex
\documentclass[10pt]{article}

\usepackage[packora]{adit_boxed_paper}
\paperpreset{Delta}

\usepackage[square,numbers,sort&compress]{natbib}
\newcommand{\lib}[1]{\textsf{#1}}  %
\usepackage{bm}
\usepackage{float}
\usepackage[section]{placeins}
\usepackage{subcaption}
\usepackage{multirow}
\usepackage{nicefrac}
\usepackage{siunitx}
\graphicspath{{figures/}}

\newcommand{\NA}{\multicolumn{1}{c}{--}}
\newcommand{\best}[1]{{\rmfamily\bfseries #1}}

\papertitle{Orbitol: Crystal Generation on Wyckoff Orbits}

\paperauthors{%
\fontsize{10.4}{12}\selectfont\raggedright\mbox{Ivan \v{Z}ugec$^{1}$}, \mbox{Joshua Rosenthal$^{2}$}, \mbox{Ivor Lon\v{c}ari\'{c}$^{3}$}, \mbox{Jan-Willem van de Meent$^{2}$}, \mbox{Tin Had\v{z}i Veljkovi\'{c}$^{2}$}
}

\paperaffiliations{%
\fontsize{9.8}{11.4}\selectfont$^1$Atomic Intelligence, Zagreb, Croatia, $^2$University of Amsterdam, $^3$Ru\dj er Bo\v{s}kovi\'{c} Institute
}

\paperabstract{%
\input{sections/abstract}
}

\paperfooter{%
\paperinfoitem{Correspondence:}{Ivan: \href{mailto:zugec.ivan@gmail.com}{zugec.ivan@gmail.com}; Tin: \href{mailto:tin.hadzi@gmail.com}{tin.hadzi@gmail.com}}
\paperinfoitem{Code and weights:}{\href{https://github.com/stases/orbitol}{github.com/stases/orbitol} $\cdot$ \href{https://huggingface.co/stases/orbitol}{huggingface.co/stases/orbitol}}
\paperinfoitem{Keywords:}{Crystal Generation, Space Groups, Wyckoff Positions, Diffusion Transformers}
}
\paperlogo{\includegraphics[height=20pt]{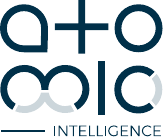}}
\hypersetup{pdftitle={Orbitol: Crystal Generation on Wyckoff Orbits}}

\begin{document}
\makepaperheader

\input{sections/introduction}
\input{sections/symmetry_models}
\input{sections/method}
\input{sections/experiments}
\input{sections/piezo}
\input{sections/conclusion}

\bibliographystyle{unsrtnat}
\bibliography{refs}

\clearpage
\appendix
\input{sections/appendix_classification}
\input{sections/appendix_implementation}
\input{sections/appendix_experiments}
\input{sections/appendix_piezo}

\end{document}

%% file: sections/abstract.tex
Generative models propose crystalline materials by learning composition and geometry, but unconditional sampling gives no direct control over the space group, which constrains whether responses such as piezoelectricity are symmetry-allowed.
We present Orbitol, a crystal generator built on the Crystalite diffusion Transformer that replaces atom tokens with Wyckoff-orbit tokens.
Given a space group and a collection of occupied Wyckoff positions, Orbitol generates the atom type and free coordinates of each orbit together with a compatible lattice, so every valid structure is invariant under the operations of the requested group.
It generates structures two to three orders of magnitude faster than the symmetry-aware baselines at the evaluated settings.
On MP-20, 13.8\% of Orbitol's attempts yielded stable or metastable, unique and novel structures that retained the requested space group after relaxation, compared with 10.8\% for SGFM, the next best model.
We also introduce SPARC, a score that combines retention of the raw detected space group with agreement between the retained-group and training distributions.
Finally, we used Orbitol to generate candidates in piezoelectric crystal classes. Of the structures selected by internal screening, 89\% passed our density functional theory checks, and 79\% also matched no Materials Project or Alexandria entry with the same space group. Their median figure of merit was nine times that of the piezoelectric materials computed in the Materials Project, and 24\% of the new candidates ranked in that set's top 10\%.

%% file: sections/introduction.tex
\section{Introduction}
\label{sec:introduction}

Generative models such as MatterGen and Crystalite propose new crystals by learning composition and geometry from known materials \citep{zeniGenerativeModelInorganic2025,HadziVeljkovic2026Crystalite}.
Many of them treat a crystal as a set of atoms, and their outputs can overrepresent low-symmetry space groups relative to the training data.
The space group P1, which contains only lattice translations \citep{Aroyo2016InternationalTables}, accounts for 1.7\% of the MP-20 training set, yet DiffCSP, MatterGen and our Crystalite reproduction produce 28.3\%, 26.4\% and 21.5\% P1 structures after relaxation.

\begin{figure}[t]
 \centering
 \includegraphics[width=\linewidth]{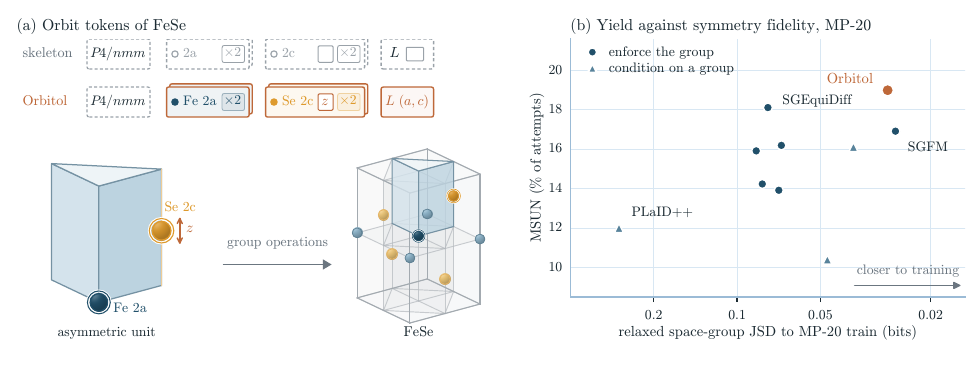}
 \caption{Orbitol fills a skeleton with orbit tokens.
 (a) Illustration for FeSe ($P4/nmm$). The skeleton fixes the space group
 and the occupied Wyckoff rows 2a and 2c. Orbitol predicts an atom type for
 each orbit, the free height $z$ of Se on row 2c, and the lattice parameters
 $a$ and $c$. Each token represents two atoms of the cell.
 The coloured piece is the standard asymmetric unit of $P4/nmm$, 1/16 of the
 cell (enlarged on the left), and contains one representative of each orbit:
 Fe at a fixed point, Se on a vertical edge at height $z$. The 16 operations of
 the group generate copies of the asymmetric unit that
 tile the cell and place the other atoms (pale).
 (b) Discovery yield on MP-20 (MSUN, the percentage of 2,500 attempts that
 are stable or metastable, unique and novel) against the Jensen--Shannon
 divergence between the space groups of the samples after unconstrained
 relaxation and those of MP-20 train (tolerance 0.1~\AA), for the ten
 symmetry-aware models of Table~\ref{tab:mp20_dng}. The Orbitol and PLaID++ points are means over three
 sampling seeds; Orbitol uses the checkpoint reported in Table~\ref{tab:mp20_dng}.}
 \label{fig:orbit_tokens}
\end{figure}

Symmetry also constrains which responses a crystal can show.
Bulk piezoelectricity is allowed only in the noncentrosymmetric crystal classes other than 432, and pyroelectricity and conventional ferroelectricity require a polar class \citep{Aroyo2016InternationalTables}.
Targeting such responses therefore requires choosing the space group of the generated structure.
Unconditional generators cannot control the space group.
A group label can guide a conditioned model, but the label alone does not guarantee that the generated structure is invariant under the group's operations \citep{zeniGenerativeModelInorganic2025}.
Other generators build the operations into generation: CrystalFormer generates Wyckoff assignments, DiffCSP++ uses supplied ones, and SymADiT and UFO-MGen model orbit representations \citep{caoSpaceGroupInformed2025,jiao2024space,Karmush2026SymADiT,Zhou2026UFOMGen}.
We study how to impose these operations during generation, and how much of the requested symmetry is retained after unconstrained relaxation.

We introduce \emph{Orbitol}, a crystal generator built on Crystalite that represents symmetry-related atoms together as \emph{Wyckoff-orbit tokens} \citep{HadziVeljkovic2026Crystalite}.
A space group and a multiset of occupied Wyckoff positions, which we call rows, form a \emph{skeleton} (Section~\ref{sec:background}).
Given a skeleton, Orbitol generates the atom type and free coordinates of each orbit together with a compatible lattice, and the decoder expands the orbits into atoms (Figure~\ref{fig:orbit_tokens}), so every valid decoded structure is invariant under the operations of the requested group.
We keep Crystalite's diffusion Transformer and adapt its geometric attention to account for all members of each orbit.

For de novo generation, Orbitol samples skeletons at their training frequencies, so we judge symmetry after unconstrained relaxation, where the group can change, and measure discovery yield with LeMat-GenBench \citep{betala2026lematgenbenchunifiedevaluationframework}.
A model can achieve high retention by concentrating its samples in a few space groups that rarely change on relaxation.
We therefore introduce SPARC, a score that combines retention of the raw detected space group with agreement between the retained-group and training distributions.

Our contributions are:
\begin{itemize}
 \item Orbitol, a diffusion model on Wyckoff-orbit tokens built on the Crystalite backbone, whose valid decoded structures are invariant under the operations of the requested space group and whose geometric attention accounts for all members of each orbit.
 \item With the sampler settings of the quality evaluation, Orbitol samples about 920 structures per second on one GPU, about 310 times faster than SGEquiDiff, the baseline with the highest observed MSUN (Section~\ref{sec:exp_cost}).
 \item On MP-20, Orbitol reaches 19.0\% stable or metastable, unique and novel yield (MSUN), the highest observed among the evaluated symmetry-aware models (18.1\% for SGEquiDiff), and 13.8\% of its attempts are MSUN structures that retain the requested group, against 10.8\% for SGFM (Table~\ref{tab:mp20_dng}).
 \item After unconstrained relaxation, Orbitol is comparable to the best evaluated baselines in symmetry fidelity, measured by the space-group Jensen--Shannon divergence to the training set and by SPARC (Section~\ref{sec:exp_symmetry}).
 \item Because it generates in a requested space group, Orbitol can target the piezoelectric crystal classes directly. Of the generated structures selected by internal screening, 89\% pass our DFT checks, and 79\% also match no MP or Alexandria entry with the same space group. Their median figure of merit is nine times that of the Materials Project's DFPT piezoelectrics, and 24\% of the new candidates rank in that set's top 10\% (Section~\ref{sec:exp_piezo}).
\end{itemize}

%% file: sections/symmetry_models.tex
\section{Related Work}
\label{sec:symmetry_models}

Crystal generators model atom types together with periodic geometry. We
compare them by what symmetry information they receive or generate and
whether their construction guarantees that the structure is invariant under
the operations of the chosen group; Table~\ref{tab:symmetry_models} and
Figure~\ref{fig:taxonomy} in Appendix~\ref{apdx:classification} classify the
models discussed here and mark those we evaluate.

\emph{Agnostic} generators specify no space group for an individual sample.
CDVAE combines a variational autoencoder with a diffusion decoder
\citep{xieCrystalDiffusionVariational}.
DiffCSP and MatterGen's base model diffuse ambient coordinates, here
fractional positions of all atoms, together with the lattice
\citep{jiaoCrystalStructurePrediction2024,zeniGenerativeModelInorganic2025}.
FlowMM generates crystals with Riemannian flow matching
\citep{millerFlowMMGeneratingMaterials2024}, and OMatG with stochastic
interpolants, which include diffusion and flow matching as special cases
\citep{hoellmerOpenMaterialsGeneration2025}.
Crystalite, on which Orbitol builds, is a diffusion Transformer on atom
tokens whose Geometry Enhancement Module turns interatomic distances into
attention biases \citep{HadziVeljkovic2026Crystalite}.
MiAD likewise requests no group \citep{Okhotin2025MiAD}. ALM Gen, the generator
of the atomistic language model ALM, decodes with MatterGen; we sampled it with
its released placeholder prompt, naming no group or composition
\citep{Edamadaka2026ALM}.

We use \emph{Conditioned} for models that receive or declare a group without
jointly constraining coordinates and lattice to make the structure invariant
under its operations. MatterGen's adapter conditions generation on a group label
\citep{zeniGenerativeModelInorganic2025}.
CrystaLLM generates crystallographic information files (CIFs)
autoregressively, token by token \citep{Antunes2024CrystaLLM}.
In the evaluated unconditional samples, CrystaLLM, CrystaLLM-$\pi$, and
PLaID++ declare their own group. CrystaLLM and CrystaLLM-$\pi$ restore its
operations in the generated CIF, allowing positions to be expanded, while
PLaID++ expands Wyckoff-labeled sites with \lib{PyXtal}
\citep{Antunes2024CrystaLLM,Bone2025CrystaLLMpi,xu2025plaidpreferencealignedlanguage}.
These models expand positions with the declared group's operations but
leave the lattice unconstrained, so the whole structure is not guaranteed to be invariant under the
operations of the declared group.

\emph{Enforced} generators construct or decode structures that are invariant under the
chosen operations. They differ in how they obtain the Wyckoff assignment:
they \emph{generate} the row identifiers, receive them as \emph{supplied}
input, or infer compatible rows \emph{implicitly} from other symmetry
information. CrystalFormer
generates one symmetry-inequivalent atom at a time: a Wyckoff letter, an atom
type and representative coordinates, which it projects onto the row
\citep{caoSpaceGroupInformed2025}.
WyFormer generates site-symmetry and enumeration identifiers that together
select a row, then obtains geometry in a separate stage
\citep{kazeevWyckoffTransformerGeneration2025}.
SGEquiDiff combines autoregressive orbit choices with group-equivariant
coordinate diffusion \citep{Chang2025SGEquiDiff}.
DiffCSP++ and SGFM instead receive the assignment as input and
constrain geometric diffusion or flow accordingly; their geometry models
do not learn the template distribution \citep{jiao2024space,Puny2025SGFM}.
SymmCD's assignment is implicit and need not identify a unique row: it
predicts site-symmetry information and projects representatives onto
compatible Wyckoff positions \citep{levy2025symmcd}.

Like DiffCSP++ and SGFM, Orbitol receives the Wyckoff assignment as input,
but it generates only the free coordinates of each orbit. CrystalFormer,
DiffCSP++ and SGFM instead generate full ambient coordinates and project them
or constrain their updates
\citep{caoSpaceGroupInformed2025,jiao2024space,Puny2025SGFM}, and GEODE
diffuses Cartesian positions with correlated orbit noise and
symmetry-preserving updates \citep{Lou2026GEODE}. Closest to Orbitol are
SymADiT, which also uses orbit tokens but evolves learned latents
\citep{Karmush2026SymADiT}, and UFO-MGen, which describes flow in a masked
free-coordinate chart \citep{Zhou2026UFOMGen};
Appendix~\ref{apdx:classification} gives details of the released UFO-MGen
and SymmCD implementations. Orbitol replaces Crystalite's atom tokens with
orbit tokens, samples skeletons from the training set and keeps them fixed
during generation (Section~\ref{sec:method_sampling}), and reduces to
Crystalite in P1 (Section~\ref{sec:method_p1}).

The detected group can be larger than the enforced one \citep{Lou2026GEODE}.
We therefore evaluate agreement with the requested group and retention of
the detected group through relaxation separately (Section~\ref{sec:exp_protocol}).

%% file: sections/method.tex
\section{Orbitol}
\label{sec:method}

\begin{figure}[t]
 \centering
 \includegraphics[width=\linewidth]{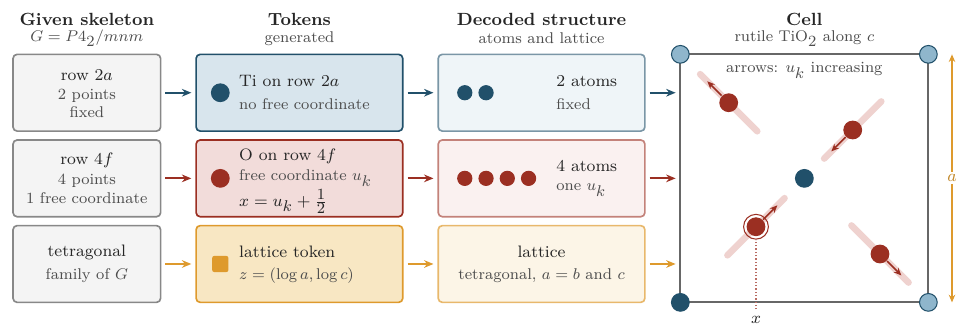}
 \caption{From a skeleton to a crystal, with rutile TiO$_2$ as the example.
 The skeleton specifies the space group $P4_2/mnm$ and the occupied rows $2a$ and $4f$. Orbitol generates one
 token for each: the atom type Ti with no free coordinate on $2a$, and the
 atom type O with one free coordinate $u_k$ on $4f$. Decoding expands these
 tokens into two fixed Ti atoms and four O atoms whose positions all depend on
 the same number $x=u_k+\tfrac12$. The lattice token specifies $a$ and $c$;
 the tetragonal family sets $b=a$ and right angles. The right panel shows the
 cell along $c$: pale lines and arrows show how the four O atoms move as $u_k$
 increases, the ringed O at $(x,x,0)$ is the representative, and pale Ti
 discs are periodic images of the Ti atom at the origin.}
 \label{fig:orbit_concepts}
\end{figure}

Orbitol keeps Crystalite's diffusion Transformer
\citep{HadziVeljkovic2026Crystalite} and uses one token per occupied orbit
instead of one per atom. Given a \emph{skeleton} consisting of a space group
$G$ and an unordered multiset of occupied Wyckoff rows, Orbitol generates an
atom type and free coordinates for each orbit, together with the free lattice
entries. Figure~\ref{fig:orbit_concepts} shows, for the example of rutile, how these
variables determine the atoms and the cell. When the skeleton is also sampled from the
training set (Section~\ref{sec:method_sampling}), Orbitol generates
composition and structure jointly, which is \emph{de novo generation} (DNG)
\citep{millerFlowMMGeneratingMaterials2024}.

\subsection{Crystals and symmetry}
\label{sec:background}

We represent an ordered periodic crystal with $N$ atoms per cell by its
lattice, fractional coordinates, and atom types
\citep{jiaoCrystalStructurePrediction2024,HadziVeljkovic2026Crystalite}:
\begin{equation*}
\begin{aligned}
\mathcal C &= (L,f,A), \qquad
L\in\mathbb R^{3\times3},\quad f\in[0,1)^{N\times3},\quad
A=(A_1,\ldots,A_N),\\
x_{i,n} &= (f_i+n)^\top L, \qquad n\in\mathbb Z^3.
\end{aligned}
\end{equation*}
Here $f_i$ is the fractional position of atom $i$ (a column vector, so
$f_i^\top$ is row $i$ of $f$), $A_i$ is its atom type, and the rows of the
nonsingular matrix $L$ are the lattice vectors. Atom $i$ also appears at the
Cartesian positions $x_{i,n}$ for every integer vector $n$; these are its
periodic images. Reordering atoms, shifting the origin, or changing the cell
basis can describe the same infinite crystal
\citep{jiaoCrystalStructurePrediction2024}.

The \emph{space group} $G$ consists of the operations that preserve the
periodic structure, including its atom types. An operation $(R,t)$ acts on
fractional coordinates as $f_i\mapsto Rf_i+t$, where $R$ and $t$ are its
matrix and translation parts. The group contains all lattice translations,
and since points that differ by one are the same point of the crystal,
finitely many operations describe how $G$ acts within one cell
\citep{Togo2024Spglib}. Applying them to a point gives its \emph{orbit}, the
set of distinct equivalent points in the cell, which we call the
\emph{members} of the orbit.

The International Tables list the \emph{Wyckoff positions} of each space
group in a table \citep{Aroyo2016InternationalTables}. We call each entry of
this table a \emph{Wyckoff row} $w$. A row describes a family of orbits by a
list of coordinate triplets that depend on its \emph{free coordinates}, and
each value of the free coordinates gives one orbit. As an example, the space
group of rutile (TiO$_2$), $P4_2/mnm$, has eleven rows, which
Table~\ref{tab:wyckoff_example} in Appendix~\ref{apdx:wyckoff_example} lists.
One of them, row $4f$, lists
\begin{equation}
 (x,x,0),\qquad (-x,-x,0),\qquad
 (-x+\tfrac12,x+\tfrac12,\tfrac12),\qquad
 (x+\tfrac12,-x+\tfrac12,\tfrac12),
 \label{eq:row_4f}
\end{equation}
Choosing $x$ places all four members of an orbit, and varying $x$ moves
them together along their own lines (Figure~\ref{fig:orbit_concepts}).

The label of a row combines its \emph{multiplicity} $m_w$, the number of
members of each orbit in the conventional cell of the tables, with a letter.
Its number of free coordinates $d_w$ says how its points can move: along a
line for $d_w=1$, within a plane for $d_w=2$, and anywhere in the cell for
$d_w=3$. Every group has exactly one row with $d_w=3$, the \emph{general
row}, whose coordinate list starts with $(x,y,z)$; in the example group it is
row $16k$. A \emph{fixed row}, with $d_w=0$, is a fixed set of points. In the
example, row $4f$ has $m_w=4$ and $d_w=1$, and rutile also places atoms on
row $2a$, with $m_w=2$, $d_w=0$ and points at $(0,0,0)$ and
$(\tfrac12,\tfrac12,\tfrac12)$.

Rows differ in the symmetry of their points. The \emph{site-symmetry group}
$S$ of a point consists of the operations of $G$ that map it exactly onto
itself. It is trivial for a generic point, that is, a point of the general
row, and larger for a point on a mirror plane, a rotation axis or an inversion center, whose orbit has $|S|$ times
fewer members than the orbit of a generic point. In the example group, an
orbit of the general row has 16 members, the points of $4f$ have $|S|=4$, and those of $2a$ have
$|S|=8$. A row collects all points whose site-symmetry groups are conjugate in
$G$, that is, $S'=gSg^{-1}$ for some $g\in G$
\citep{Aroyo2016InternationalTables}.

A crystal places its atoms on some of these rows, its \emph{occupied} rows.
Several orbits can occupy a row with free coordinates, each with its own
coordinate values, so the row can appear several times in a skeleton. The
orbits must be distinct, and different values can give the same orbit, as $x$
and $-x$ do on row $4f$ of the example. A fixed row is a single orbit and appears at most once.
All atoms of an orbit have the same atom type, because the operations of $G$
map each atom onto an atom of the same type. In crystallographic terms, an
occupied orbit is one symmetry-independent atom, one line in the atom-site
list of a symmetrized CIF, and a skeleton is a space group with its
\emph{Wyckoff sequence}, the letters of its occupied rows with repeats counted
\citep{Allmann2007ICSD}. Appendix~\ref{apdx:wyckoff_example} works through
two examples, rutile and CrF$_4$, which occupy different rows of the same
group.

\subsection{Orbit and lattice representation}
\label{sec:method_orbits}
\label{sec:method_lattice}

Orbitol generates only the variables that the skeleton leaves free. For a
skeleton with $K$ occupied orbits on rows $w_1,\ldots,w_K$, which may repeat,
it uses one token per orbit and one lattice token
(Figure~\ref{fig:orbit_tokens} shows an example for FeSe). For orbit $k$,
Orbitol generates an atom type $A_k$ and free coordinates
$u_k\in[-\tfrac12,\tfrac12)^{d_{w_k}}$; adding $\tfrac12$ gives the coordinates
used by the tables.

Each entry of a row's coordinate list, which we take from PyXtal
\citep{Fredericks2021PyXtal}, is an affine function of the free coordinates.
We write entry $j$ of row $w$ as a \emph{member map} $(W_{w,j},v_{w,j})$, with
a $3\times d_w$ matrix $W_{w,j}$ and an offset $v_{w,j}$, so that the members
of orbit $k$ are
\begin{equation}
 f_{kj}=\mathrm{mod}_1\!\left(W_{w_k,j}(u_k+\tfrac12)+v_{w_k,j}\right),
 \qquad j=1,\ldots,m_{w_k},
 \label{eq:orbit_members}
\end{equation}
where $f_{kj}$ is the fractional position of atom $j$ of orbit $k$ and
$\mathrm{mod}_1$ wraps each coordinate into $[0,1)$.

We call the first member $f_{k1}$ the \emph{representative} and its map
$(J_w,b_w)=(W_{w,1},v_{w,1})$ the \emph{chart} of row $w$: the columns of $J_w$
are the directions in which the representative can move, and $b_w$ is its
offset. Every other member is the image of the representative under an
operation $(R_j,t_j)$ of $G$, so $W_{w,j}=R_jJ_w$ and
$v_{w,j}=\mathrm{mod}_1(R_jb_w+t_j)$. Changing $u_k$ therefore moves all
members together and keeps the symmetry.
Orbitol outputs primitive cells, with $m^{\mathrm p}_w$ atoms per orbit;
Appendix~\ref{apdx:implementation_tables} explains the conversion from the
conventional cells of the tables and the equivalent descriptions with which
we augment the training crystals.

The lattice metric must be invariant under the point-group operations, so its
allowed lengths and angles depend on the crystal family of $G$ (triclinic,
monoclinic, orthorhombic, tetragonal, hexagonal or cubic) and on the cell
setting \citep{Aroyo2016InternationalTables}. Crystalite describes the cell by
a latent $y\in\mathbb R^6$, the logarithms of the three diagonal entries and
the three off-diagonal entries of a lower-triangular matrix whose rows are the
lattice vectors; Orbitol builds the conventional cell in this way. The crystal
family fixes some of these entries and ties others together. Orbitol
generates only the free entries $z$ and reconstructs $y$ with a
\emph{crystal-family chart},
\begin{equation}
 y=\ell_G(z),
 \label{eq:lattice_chart}
\end{equation}
which fills in all fixed and dependent entries
(Appendix~\ref{apdx:implementation_lattice}).

As a worked example, the skeleton of rutile selects $w_1=2a$ and $w_2=4f$, so
$K=2$ (Figure~\ref{fig:orbit_concepts}). Orbitol generates $A_1=\mathrm{Ti}$
for the first orbit, which has no free coordinate, and $A_2=\mathrm O$ and one
number $u_2$ for the second, which places the four O atoms of
Equation~\eqref{eq:row_4f} at $x=u_2+\tfrac12$. The first entry, $(x,x,0)$, is
the representative, so $J_{4f}=(1,1,0)^\top$ and $b_{4f}=0$; the third entry
has $W_{4f,3}=(-1,1,0)^\top$ and $v_{4f,3}=(\tfrac12,\tfrac12,\tfrac12)^\top$.
The two Ti atoms stay at fixed positions. The tetragonal family sets $b=a$ and right angles, so the lattice
token specifies $z=(\log a,\log c)$, from which the chart reconstructs
$y=(\log a,0,\log a,0,0,\log c)$. Two atom types and three numbers, $u_2$,
$\log a$ and $\log c$, thus determine all six atoms and the cell.

\subsection{Diffusion formulation}
\label{sec:method_diffusion}

For a given skeleton, the free variables form the state $(h,u,z)$:
Crystalite's 16-dimensional atom-type encoding $h_k$ of each $A_k$,
the free coordinates $u=(u_1,\ldots,u_K)$ and the free lattice entries $z$.
The skeleton itself is not diffused; it fixes the member maps and charts
that turn a state into a crystal.

Noise of a given size should move atoms by a similar amount on every row,
but free coordinates are not lengths. On row $4f$ of rutile, for example, a
step $\delta$ in $x$ moves each O atom along $(1,1,0)$, by $\sqrt2|\delta|$
in fractional coordinates. We therefore measure a step $\delta$ of the free
coordinates by the displacement of the atoms,
\begin{equation}
 \|J_w\delta\|_{M_G}^2=\delta^\top H_w\delta,\qquad
 H_w=J_w^\top M_GJ_w,\qquad
 M_G=\frac{1}{|\mathcal R_G|}\sum_{R\in\mathcal R_G}R^\top R,
 \label{eq:chart_metric}
\end{equation}
where $\|v\|_{M_G}^2=v^\top M_Gv$ and $\mathcal R_G$ contains the
point-group rotations of $G$. Averaging over the rotations gives every member
of an orbit the same displacement, also in hexagonal axes, where the
rotation matrices are not orthogonal (Appendix~\ref{apdx:chart_metric}).
The metric is fixed, so the coordinate noise does not depend on the
lattice, which is itself generated. Following Crystalite's EDM formulation
\citep{NEURIPS2022_a98846e9,HadziVeljkovic2026Crystalite}, one noise level
$\sigma$ per structure perturbs the three channels jointly,
\begin{equation}
 h_k^\sigma=h_k+\sigma\epsilon_{h,k},\qquad
 u_k^\sigma=\mathrm{wrap}\big(u_k+\sigma\xi_k\big),\qquad
 z^\sigma=z+\sigma\epsilon_z,
 \label{eq:orbit_noising}
\end{equation}
with standard Gaussian draws $\epsilon$, coordinate noise
$\xi_k\sim\mathcal N(0,H_{w_k}^{-1})$, and $\mathrm{wrap}(v)=v-\mathrm{round}(v)$.
The covariance $H_w^{-1}$ makes the atom displacements equally large in every
allowed direction. On row $4f$ of rutile, $H_{4f}=2$, so the noise on $u_k$
has standard deviation $\sigma/\sqrt2$. Fixed rows receive no coordinate
noise.

\subsection{Tokens and network}
\label{sec:method_model}
\label{sec:method_conditioning}
\label{sec:method_inputs}

The network is Crystalite's Transformer applied to one token per orbit,
followed by one lattice token. As in Crystalite, each token sums the
embeddings of its noisy channels. For the position, an orbit token embeds
only its representative: the row determines every other member from it, so
their positions add no information. The lattice token embeds the noisy
lattice in Crystalite's six-entry form $y^\sigma=\ell_G(z^\sigma)$, with the
fixed and tied entries filled in by the chart:
\begin{equation}
 \begin{aligned}
 \eta_k&=E_h(c_{\mathrm{in},h}h_k^\sigma)+E_x(u_k^\sigma,w_k)
 +e_{\mathrm{orb}}+E_w(w_k),\\
 \eta_L&=E_L\big(c_{\mathrm{in},z}\,\ell_G(z^\sigma)\big)+e_{\mathrm{lat}}.
 \end{aligned}
 \label{eq:token_inputs}
\end{equation}
Here $E_h$, $E_x$, $E_w$ and $E_L$ embed the type features, position, row and
lattice, $c_{\mathrm{in},h}$ and $c_{\mathrm{in},z}$ are EDM input scales, and
the learned vectors $e_{\mathrm{orb}}$ and $e_{\mathrm{lat}}$ mark orbit and
lattice tokens. The position embedding
$E_x(u,w)=E_F\big(\mathrm{mod}_1(J_w(u+\tfrac12)+b_w)\big)$ applies
Crystalite's Fourier embedder $E_F$ to the representative $f_{k1}$ of
Equation~\eqref{eq:orbit_members} at the noisy free coordinates; on a fixed
row, this is the fixed position $b_w$. Appendix~\ref{apdx:implementation_inputs}
describes a variant that embeds the free coordinates themselves.

The same free coordinate describes different motions on different rows, so
the network needs each token's row and the space group. The row embedding
$E_w$ gives each token its row. For the group, we extend Crystalite's noise
conditioning: Crystalite embeds the noise coordinate
$c_{\mathrm{noise}}=\tfrac14\log\sigma$ with a small MLP $E_\sigma$ and
injects it into every block through adaptive layer normalization (AdaLN)
\citep{peeblesScalableDiffusionModels2023}, and Orbitol adds a group
embedding,
\begin{equation}
 c=E_\sigma(c_{\mathrm{noise}})+E_G(G).
 \label{eq:group_conditioning}
\end{equation}
Each row or group embedding sums a learned lookup vector and an MLP
embedding of crystallographic descriptors, such as the multiplicity, the
number of free coordinates and the chart geometry of a row, or the crystal
system, centering and point-group operations of a group
(Appendix~\ref{apdx:implementation_conditioning}). These embeddings tell the
network which group and rows it is filling. The symmetry does not depend on
them: Equation~\eqref{eq:orbit_members} and the lattice chart give every
valid decoded structure the symmetry of $G$, whatever the network predicts.

After the last block, output heads turn the state of each orbit token into
raw predictions $F_{h,k}$ and $F_{u,k}$, and the state of the lattice token
into $F_z$. Like the input, the coordinate prediction concerns the
representative: one head, shared by all rows, predicts a three-dimensional
update of its position. The representative can move only along the columns
of $J_w$, so we take the allowed update closest to the prediction under
$M_G$:
\begin{equation}
 \begin{aligned}
 \widehat F_{u,k}
 &=\operatorname*{arg\,min}_{\delta\in\mathbb R^{d_{w_k}}}
 \|J_{w_k}\delta-F_{u,k}\|_{M_G}^2=\Pi_{w_k}F_{u,k},\\
 \Pi_w&=H_w^{-1}J_w^\top M_G.
 \end{aligned}
 \label{eq:orbit_projection}
\end{equation}
On the general row, the representative $(x,y,z)$ can move in every
direction, so $J_w=I$ and $\Pi_w=I$; in P1, where every orbit lies on the
general row, this is Crystalite's coordinate head. The lattice head predicts
the six entries of $y$, and its readout $\Gamma_G$ averages the predictions of
tied entries and masks the fixed ones. With the EDM coefficients
$c_{\mathrm{skip}}$ and $c_{\mathrm{out}}$
(Appendix~\ref{apdx:implementation_noise}), the denoised predictions are
\begin{equation}
 \begin{aligned}
 D_{h,k}&=c_{\mathrm{skip},h}h_k^\sigma+c_{\mathrm{out},h}F_{h,k},\\
 D_{u,k}&=s_u(u_k^\sigma)+c_{\mathrm{out},u}\widehat F_{u,k},\\
 D_z&=c_{\mathrm{skip},z}z^\sigma+c_{\mathrm{out},z}\Gamma_GF_z,
 \end{aligned}
 \label{eq:orbit_denoisers}
\end{equation}
where $s_u$ is the coordinate skip of Section~\ref{sec:method_targets}.

\subsection{Denoising targets}
\label{sec:method_targets}

The type and lattice channels keep Crystalite's EDM-weighted squared errors.
The coordinate channel needs care because free coordinates are periodic: a
clean point $u$ and its images $u+n$, $n\in\mathbb Z^{d_w}$, describe the same
orbit. The sampler needs the denoiser to predict the posterior mean of the
clean coordinates given the noisy ones. Crystalite's \emph{wrapped-residual
objective} (M) compares a prediction with whichever image of the clean point
is nearest to it, and its squared-error optimum need not be this mean
(Appendix~\ref{apdx:implementation_objectives}). For example, if the clean
coordinate is $u=-\tfrac12$ and the noisy one is $u^\sigma=0$, the images
$-\tfrac12$ and $\tfrac12$ are equally likely and their mean is $0$, whereas M
is minimized by predicting either image.
The \emph{periodic-posterior target} (P) averages the images, weighted by
their Gaussian probability given the noisy point:
\begin{equation}
 D^\star(u^\sigma,u)=\sum_{n\in\mathbb Z^{d_w}}\pi_n\,(u+n),\qquad
 \pi_n\propto\exp\!\left[-\frac{\|u^\sigma-u-n\|_{H_w}^2}{2\sigma^2}\right],
 \label{eq:orbit_periodic_target}
\end{equation}
where $\|v\|_{H_w}^2=v^\top H_wv$ and $\sum_n\pi_n=1$. At low noise, $D^\star$
is the image nearest to $u^\sigma$; at high noise, it approaches $u^\sigma$
without preferring a chart origin. For a given clean point, $D^\star$ is the
posterior mean of its images, and regression onto it averages this mean
over the clean points compatible with the noisy state. The coordinate loss
is the EDM-weighted squared $H_w$ error between $D_{u,k}$ and $D^\star$
(Appendix~\ref{apdx:implementation_objectives}). Target P is a choice of
training objective, not part of the symmetry construction: Orbitol supports
both, and Section~\ref{sec:method_p1} uses M to reproduce Crystalite exactly
in P1. We train the reported models with P because its regression optimum is
the posterior mean that the sampler assumes.

The skip term should respect the same periodicity: shifting the noisy input
by an integer $n$ should shift the prediction by $n$. We therefore replace the
EDM skip $c_{\mathrm{skip},u}u$ by the sine skip
$s_u(u)=u-(1-c_{\mathrm{skip},u})\sin(2\pi u)/(2\pi)$, applied componentwise,
which satisfies $s_u(u+n)=s_u(u)+n$ and agrees with the EDM skip to first
order at $u=0$. Appendix~\ref{apdx:ablations} compares the two skips.
DiffCSP's wrapped-normal diffusion and GEODE also use periodic score targets
\citep{jiaoCrystalStructurePrediction2024,Lou2026GEODE}; Orbitol applies such
a target to the free coordinates of each row, measured with $H_w$.

\subsection{Geometry in orbit attention}
\label{sec:method_gem}

\begin{figure}[t]
 \centering
 \includegraphics[width=0.95\linewidth]{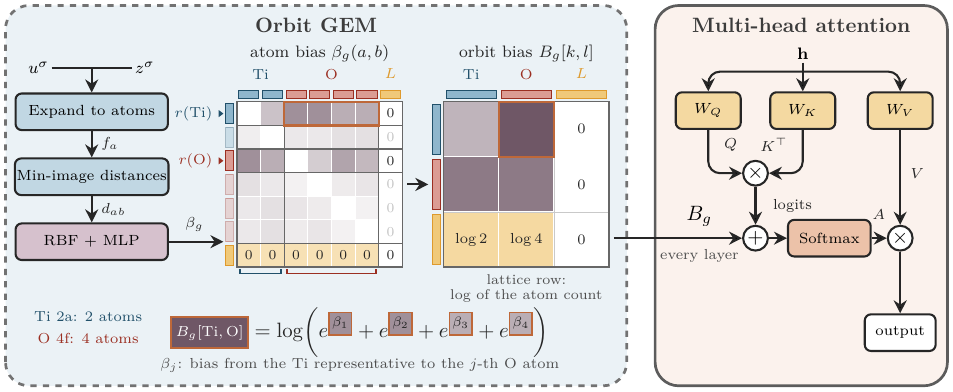}
 \caption{GEM for orbit tokens, with rutile as the example (TiO$_2$,
 $P4_2/mnm$; two Ti and four O atoms per cell). GEM turns interatomic distances into attention
 biases, but Orbitol has one token per orbit, so the atom-pair biases
 $\beta_g(a,b)$ are combined into one bias per pair of orbit tokens. The atoms
 of an orbit are related by the group operations and have the same set of
 distances to every other orbit, so one representative row $r(k)$ per orbit
 is kept (pale rows are dropped). The atoms of a key orbit share one key and
 value, so we sum their attention weights:
 $B_g[k,l]=\log\sum_{b\in l}\exp\beta_g\big(r(k),b\big)$ gives the orbit the
 total weight of its atoms, as shown at the bottom for the Ti representative
 and the four O atoms. The lattice token has zero atom biases, so its row
 of $B_g$ is the log atom count of each orbit. The orbit biases are added to the
 attention logits in every layer.}
 \label{fig:gem_orbits}
\end{figure}

Crystalite's Geometry Enhancement Module (GEM) adds a geometry-dependent
bias to the attention scores between atom tokens.
In Orbitol, each token is an orbit that corresponds to several atoms, but
the network should still see the geometry of all of them, so we contract the
atom-pair biases into one bias per pair of orbit tokens
(Figure~\ref{fig:gem_orbits}). We expand each noisy orbit
into its conventional members with Equation~\eqref{eq:orbit_members}, convert them to
primitive-cell members with Equation~\eqref{eq:orbit_primitive}, and approximate
minimum-image distances using a finite search in a reduced cell basis.
For head $g$, a distance-only MLP on radial
features of absolute and volume-normalized distances gives the atom bias
$\beta_g(a,b)$, with $\beta_g(a,a)=0$. With exact distances in a cell that
is compatible with the group, all atoms in a query orbit have the same distances
to each key orbit up to a permutation, so we keep one representative row
$r(k)$ per query orbit. The atoms of key orbit $l$ share a key and value;
we add their attention weights by taking the log-sum-exp of their biases:
\begin{equation}
 B_g[k,l]=\log\sum_{b\in l}\exp\beta_g\big(r(k),b\big).
 \label{eq:orbit_gem_aggregation}
\end{equation}
Since $\exp B_g[k,l]=\sum_{b\in l}\exp\beta_g(r(k),b)$, this gives the same
attention output for the selected query as an expanded layer whose members
share each orbit's key and value. Averaging the atom biases would change the
total attention assigned to the orbit. The lattice query row
receives $\log m^{\mathrm p}_{w_l}$, the log atom count of each key orbit,
and the lattice key column receives zero. In P1, the sum has one term, so
orbit GEM becomes an atom-level GEM, as in Crystalite.
Appendix~\ref{apdx:implementation_gem} works through rutile as an example
and gives the distance implementation.

\subsection{Sampling}
\label{sec:method_sampling}

Generation first samples a skeleton from the empirical distribution of
training skeletons, so every generated structure shares its skeleton with
a training structure. MP-20 has 3,290 distinct training
skeletons; 5.2\% of validation structures and 4.9\% of test structures have
skeletons absent from the training set. Given the skeleton, Crystalite's EDM
sampler with stochastic churn and Heun correction integrates the three
channels from noise with the denoisers of
Equation~\eqref{eq:orbit_denoisers}.
Appendix~\ref{apdx:sampler_sweep} gives the settings of each experiment. We then
decode the atom types, place every atom of each orbit with
Equation~\eqref{eq:orbit_members}, reconstruct the lattice with $\ell_G$, and
convert the resulting structure to a primitive cell
(Appendix~\ref{apdx:implementation_sampling}).

\subsection{Equivalence to Crystalite in P1}
\label{sec:method_p1}

In P1, every orbit is a single atom with three free coordinates, and the
lattice is unconstrained. The chart, coordinate metric and projection
become identities, and each orbit token becomes an atom token, so by
construction Orbitol has Crystalite's representation and network. With matched
type encodings, weights, execution modes and sampler settings, objective M,
GEM disabled, and zero row and group embeddings, the P1 implementation
reproduces Crystalite's denoiser, loss and sampler exactly
(Appendix~\ref{apdx:implementation_parity}). Our reported configurations
instead use the periodic-posterior target with the sine skip
(Section~\ref{sec:method_targets}), a distance-only GEM without
Crystalite's displacement features and noise gate
(Appendix~\ref{apdx:implementation_gem}), and learned row and group
embeddings, which are constant in P1.

%% file: sections/experiments.tex
\section{Experiments}
\label{sec:experiments}

We evaluate de novo generation on MP-20 and Alex-MP-20 by discovery yield,
symmetry after relaxation, and sampling cost.
Section~\ref{sec:exp_piezo} then applies Orbitol to piezoelectric candidates.

\subsection{Setup}
\label{sec:exp_setup}

We trained on the training splits of MP-20 and Alex-MP-20. MP-20 contains
Materials Project structures; Alex-MP-20 combines Materials Project and
Alexandria data. Both restrict cells to at most 20 atoms
\citep{xieCrystalDiffusionVariational,zeniGenerativeModelInorganic2025}.
We sampled untyped skeletons at their training frequencies and generated
atom types, free coordinates and lattice conditional on them.
On MP-20, we report the Orbitol checkpoint with the highest observed MSUN in
the generation evaluations during training; on Alex-MP-20, the checkpoint at
650k steps.
Both use Crystalite's EDM sampler
\citep{NEURIPS2022_a98846e9,HadziVeljkovic2026Crystalite}
(Appendix~\ref{apdx:sampler_sweep}). We compare with the models of
Section~\ref{sec:symmetry_models}, using released checkpoints where available,
our train-only SymmCD and WyFormer retrainings
(WyFormer uses DiffCSP++ for geometry) and our Crystalite reproduction
(Appendix~\ref{apdx:experiments}).

We evaluate the first 2,500 attempts of each model; generation, decoding and
structure-check failures stay in the denominator, and nothing is replaced.
For PLaID++, each language-model generation is an attempt; its released
sampler discards generations that fail to decode and samples again; we
count such generations as failures. We pre-relax
all structures with NequIP-OAM-L \citep{Kavanagh2026NequIPOAM} for 500 FIRE
steps with a Frechet cell filter and no symmetry constraint on positions or
cell. LeMat-GenBench then scores energies with ORB, MACE and UMA against
their hulls and novelty against LeMat-Bulk
\citep{betala2026lematgenbenchunifiedevaluationframework}, using the raw
geometry where relaxation failed. We detect symmetry with \lib{spglib}
through \lib{pymatgen}'s \texttt{SpacegroupAnalyzer} at a positional
tolerance of 0.1~\AA\ and an angular tolerance of $5^\circ$, for generated
and training structures alike. Each model is one training run and one set
of 2,500 samples; for the MP-20 Orbitol model and PLaID++, we average three
sampling seeds.

\subsection{Evaluation metrics}
\label{sec:exp_protocol}

\emph{Discovery metrics} are shares of all 2,500 attempts. SUN and MSUN count
valid structures that are unique among the model's attempts and novel
against LeMat-Bulk, with predicted energy above the hull at most 0 (SUN,
stable) or at most 0.1~eV/atom (MSUN, stable or metastable). Joint yield
counts attempts that satisfy MSUN and retain the requested or declared space
group after relaxation.

\emph{Symmetry fidelity} metrics ask whether relaxed structures keep their
groups and whether these groups match the training distribution. P1 is the
share of relaxed structures detected in P1, and JSD is the Jensen--Shannon
divergence, in bits, between their space groups and those of the training
set. Retention is the share of all attempts whose relaxed structure has
exactly the requested group, or the declared group for models that declare
one, so relaxing into a supergroup counts as a loss. Failed relaxations and
implausible endpoints are not \emph{eligible}: they are left out of P1 and
JSD and count as losses for retention, joint yield and SPARC
(Appendix~\ref{apdx:evaluation}). Appendix Table~\ref{tab:exp_metrics}
collects the definitions.

Retention and coverage capture different failures. A model can retain nearly
every group by generating only a few groups that rarely change on
relaxation, such as rock-salt structures in $Fm\bar{3}m$, while covering
little of the training distribution. Conversely, a model that requests groups
at their training frequencies can still lose many of them on relaxation:
38--48\% for the enforced baselines other than SGFM (Table~\ref{tab:mp20_dng}).
SPARC (symmetry preservation and reference coverage) requires both, as a
product:
\begin{equation}
\mathrm{SPARC} = 100\,R_{\rm raw}\,\mathrm{BC}(p,q)^2,
\label{eq:sparc}
\end{equation}
where $R_{\rm raw}$ is the share of all attempts whose eligible relaxed
structure keeps the group detected in its raw structure, and
$\mathrm{BC}(p,q)=\sum_g\sqrt{p_g q_g}$ is the Bhattacharyya coefficient
between the training frequencies $p_g$ and the frequencies $q_g$ of the
groups among the retained attempts. SPARC is thus the percentage of attempts
that keep their group, discounted by how far their groups are from the
training distribution: it reaches 100 only if every attempt keeps its group
and the kept groups follow the training distribution, and it is zero if no
attempt keeps its group. Because it uses the raw detected group rather than a
requested one, SPARC also scores agnostic models. Figure~\ref{fig:sparc}(b)
plots the two factors, and Appendix~\ref{apdx:sparc} gives the derivation and
the score of an ideal sampler.

\subsection{Discovery yield}
\label{sec:exp_dng}

\input{tables/v3/mp20_dng}

On MP-20, Orbitol reached 19.0\% MSUN and 13.8\% joint yield, the highest
observed among the evaluated symmetry-aware models
(Table~\ref{tab:mp20_dng}); SGEquiDiff was next in MSUN (18.1\%) and SGFM in
joint yield (10.8\%). We scored the agnostic models for symmetry only
(Tables~\ref{tab:full_results} and~\ref{tab:full_results_symmetry}). The
LeMat-GenBench leaderboard in Appendix Table~\ref{tab:lemat_metrics} lists
values from the authors' own submissions
\citep{betala2026lematgenbenchunifiedevaluationframework}; for the
symmetry-aware models in both, our measured MSUN is 3.4--13.4 percentage
points higher, so the two are not directly comparable.

\label{sec:exp_alex}
On Alex-MP-20, Orbitol reached 34.2\% MSUN and retained its requested group
in 71.6\% of attempts (Appendix Tables~\ref{tab:full_results}
and~\ref{tab:full_results_symmetry}). GEODE reports yields without relaxation
and after relaxation with a different potential, and its MSUN excludes SUN \citep{Lou2026GEODE}; Appendix
Table~\ref{tab:lemat_metrics_alex} lists them next to ours. LeMat-Bulk
contains Alexandria, so novelty on this dataset is measured against a
reference that overlaps the training source.

\subsection{Symmetry after relaxation}
\label{sec:exp_symmetry}

\begin{figure}[t]
    \centering
    \includegraphics[width=\linewidth]{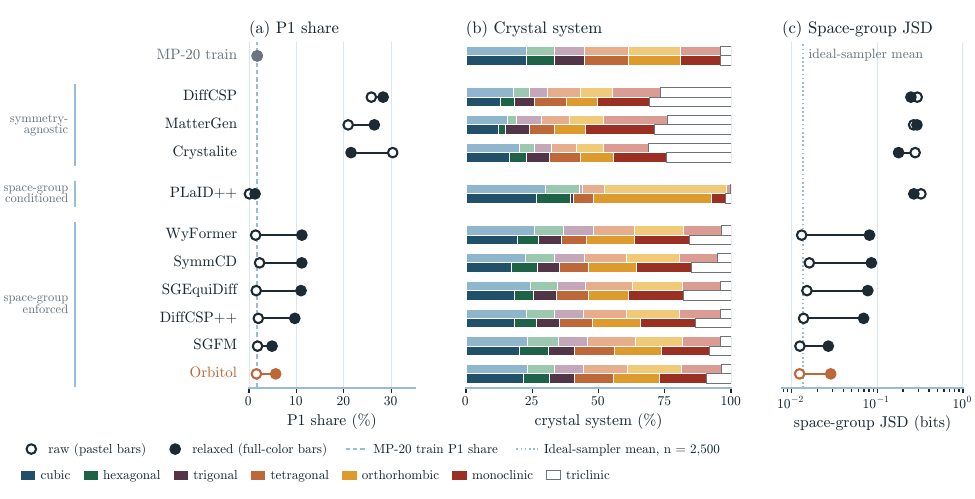}
    \caption{Raw and relaxed symmetry on MP-20 at a positional detection
    tolerance of 0.1~\AA.
    Hollow markers and pastel bars show raw outputs; filled markers and
    full-color bars show endpoints after 500 unconstrained FIRE steps.
    Panels show (a) P1 share on a linear axis, (b) crystal-system shares
    and (c) space-group JSD in bits on a logarithmic axis. The dashed line
    in (a) marks the P1 share of MP-20 train; the raw P1 share of PLaID++
    (0.1\%) is near zero. The dotted line in (c) marks the
    mean JSD of an ideal sampler of the training distribution with 2,500
    draws: 0.0137 bits, with a 5th--95th percentile range of 0.0114--0.0160.
    Crystalite is our reproduction; SymmCD and WyFormer are retrained.
    Orbitol and PLaID++ are means over three sampling seeds.}
    \label{fig:symmetry_distribution}
\end{figure}

The agnostic models DiffCSP, MatterGen and our Crystalite reproduction
overproduce P1 after relaxation (Figure~\ref{fig:symmetry_distribution}).
Among the enforced models, Orbitol and SGFM have the lowest relaxed P1
shares and JSDs, 0.029 and 0.027 bits; the difference is within the
finite-sample spread of an ideal sampler, so more sampling seeds would be
needed to resolve it. Orbitol retained its requested group in 74.7\% of
attempts, close to SGFM and above the other enforced models
(Table~\ref{tab:mp20_dng}).

Retention and a low P1 share alone do not show agreement with the training
distribution. PLaID++ retained its declared group in 76.6\% of attempts,
about as often as Orbitol (74.7\%), and had a low P1 share, yet its relaxed
JSD was 0.267 bits, because its outputs concentrate on
a few groups, such as $Pnma$ and $Fm\bar{3}m$, that often survive relaxation.

\begin{figure}[t]
    \centering
    \includegraphics[width=\linewidth]{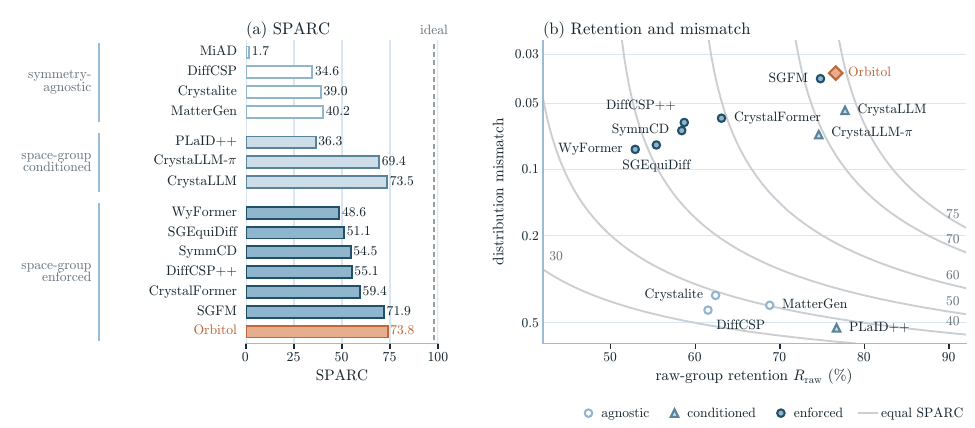}
    \caption{SPARC on MP-20. (a) SPARC by model
    (Equation~\ref{eq:sparc}); the dashed line marks the average score of
    an ideal sampler at 2,500 attempts (97.9). (b) The two factors of SPARC:
    the share of attempts whose relaxed space group equals the group
    detected in the raw output, and the distribution mismatch, one minus
    the squared Bhattacharyya coefficient between the groups of the retained
    structures and the training distribution (logarithmic axis, better
    toward the top). Curves connect points of equal SPARC; each is labeled
    once with its value. MiAD ($R_{\rm raw}=12\%$) lies outside the axes.}
    \label{fig:sparc}
\end{figure}
\begin{figure}[t]
    \centering
    \includegraphics[width=\linewidth]{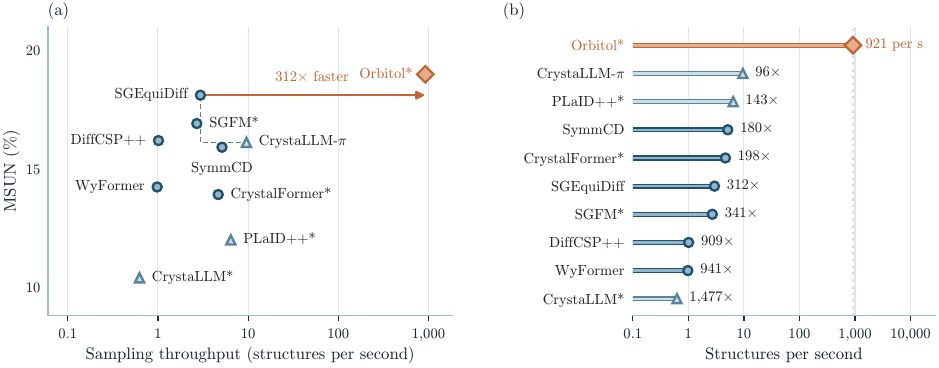}
    \caption{Sampling cost on MP-20 on one RTX 6000 Ada. (a) MSUN from
    Table~\ref{tab:mp20_dng} against sampling throughput; the dashed line is
    the frontier of the baselines. (b) Throughput, with Orbitol's speedup
    over each baseline. Timings, measured after warm-up, run from the
    prior draw to decoded structures and count failed attempts. Each model
    uses its released sampler and the batch size with the highest measured
    throughput. Sampler settings match the quality evaluation.
    Orbitol uses the sampler of Table~\ref{tab:mp20_dng}
    with 25 Heun steps. $^{*}$Measurement notes in
    Appendix~\ref{apdx:cost}.}
    \label{fig:cost}
\end{figure}

SPARC separates these cases (Figure~\ref{fig:sparc}). On MP-20, Orbitol
scores 73.8, close to CrystaLLM and SGFM and above all other models, whereas
PLaID++ scores 36.3; Alex-MP-20 scores are in
Table~\ref{tab:full_results_symmetry}. These comparisons depend on the
relaxer and the detector: retention differences of 1--3 percentage points
can fall within run-to-run relaxation variation
(Appendix~\ref{apdx:additional_results}), which also repeats the comparison
at 0.01~\AA\ (Figure~\ref{fig:symmetry_distribution_0p01}).
Appendices~\ref{apdx:evaluation} and~\ref{apdx:sparc} give alternative
retention rates and a requested-group variant of SPARC.

\subsection{Sampling cost}
\label{sec:exp_cost}

We timed every model of Table~\ref{tab:mp20_dng} on one RTX 6000 Ada with
its released code and the sampler settings and precision of the quality
evaluation. Some released samplers were slow
for reasons unrelated to the model, such as SGFM's unbatched sampling loop,
so we batched them and tuned each batch size without changing the samples or
step counts. Timing runs from the prior draw to decoded structures, counts
failed attempts, and excludes model loading, compilation and relaxation
(Appendix~\ref{apdx:cost}). With the sampler of Table~\ref{tab:mp20_dng},
Orbitol samples about 920 structures per second in large batches, about 310
times faster than SGEquiDiff, the baseline with the highest observed MSUN,
and 96 to 1,477 times faster than the other symmetry-aware baselines
(Figure~\ref{fig:cost}). The speed comes from a short sampler (25 Heun
steps) and an efficient implementation: one token per orbit shortens the
sequences, crystals are packed without padding, the network runs as a
compiled CUDA graph, and decoding is vectorized (Appendix~\ref{apdx:speed}).

%% file: tables/v3/mp20_dng.tex
\begin{table*}[!b]
\centering
\small
\caption{De novo generation on MP-20. Values are from 2,500 sampled attempts per model; the MP-20 Orbitol and PLaID++ rows are means over three sampling seeds, each metric averaged per seed. Metrics are defined in Section~\ref{sec:exp_protocol}. Training P1: $1.7\%$. Joint yield counts attempts that satisfy MSUN and retain the requested or declared group after relaxation, with symmetry detected at 0.1~\AA. Bold marks maxima for SUN, MSUN, joint yield, requested group retained and SPARC, and the minimum JSD. Alex-MP-20 results and all models with marginal diagnostics are in Tables~\ref{tab:full_results} and~\ref{tab:full_results_symmetry}.}
\label{tab:mp20_dng}
\setlength{\tabcolsep}{5pt}
\begin{tabular}{@{}l rrr rrrr@{}}
\toprule
& \multicolumn{3}{c}{Discovery (\%)} & \multicolumn{4}{c}{After relaxation} \\
\cmidrule(lr){2-4}\cmidrule(l){5-8}
Model & SUN & MSUN & \shortstack{Joint\\yield} & \shortstack{Requested group\\retained (\%)} & SPARC & P1 (\%) & JSD (bits) \\
\midrule
\multicolumn{8}{@{}l}{\emph{Space group conditioned}} \\
CrystaLLM & 1.00 & 10.4 & 4.6 & \best{77.8}$^{d}$ & 73.5 & 4.7 & 0.047 \\
CrystaLLM-$\pi$ & \best{1.40} & 16.1 & 8.8 & 74.3$^{d}$ & 69.4 & 4.6 & 0.038 \\
PLaID++$^{a}$ & 0.95 & 12.0 & 7.2 & 76.6$^{d}$ & 36.3 & 1.3 & 0.267 \\
\addlinespace
\multicolumn{8}{@{}l}{\emph{Space group enforced}} \\
CrystalFormer & 0.80 & 13.9 & 7.8 & 62.3 & 59.4 & 11.1 & 0.071 \\
SGEquiDiff & 0.68 & 18.1 & 8.1 & 54.7 & 51.1 & 11.0 & 0.077 \\
WyFormer$^{b}$ & 0.40 & 14.2 & 6.2 & 51.8 & 48.6 & 11.2 & 0.081 \\
SymmCD$^{b}$ & 0.84 & 15.9 & 9.0 & 57.4 & 54.5 & 11.1 & 0.085 \\
DiffCSP++ & 0.60 & 16.2 & 7.6 & 57.3 & 55.1 & 9.7 & 0.069 \\
SGFM$^{c}$ & 1.00 & 16.9 & 10.8 & 73.4 & 71.9 & 4.9 & \best{0.027} \\
\addlinespace
\multicolumn{8}{@{}l}{\emph{Ours (space group enforced)}} \\
Orbitol & 1.11 & \best{19.0} & \best{13.8} & 74.7 & \best{73.8} & 5.6 & 0.029 \\
\bottomrule
\end{tabular}
\par\smallskip
\begin{minipage}{\textwidth}
\footnotesize
Orbitol uses the checkpoint selected by the highest observed MSUN in generation evaluations conducted during training; sampler settings are in Appendix~\ref{apdx:sampler_sweep}.
An ideal sampler of the training distribution with 2,500 draws has a JSD of 0.0137~bits on average (5th--95th percentile range 0.0114--0.0160; Figure~\ref{fig:symmetry_distribution}). The Orbitol--SGFM JSD difference is small relative to this finite-sample spread; additional samples would be needed to resolve it.
$^{a}$Released DPO iteration-7 adapter with the authors' sampling settings (4-bit base, temperature 1.3, top-p 0.9, batch 1,024); generations that fail to decode count as failed attempts.
$^{b}$Our retraining on the training split only; WyFormer geometry from DiffCSP++.
$^{c}$Wyckoff templates drawn from the training set, as for Orbitol.
$^{d}$Uses the group declared by the model.
\end{minipage}
\end{table*}

%% file: sections/piezo.tex
\makeatletter\let\orbitol@secFB\@fb@secFB\let\@fb@secFB\relax\makeatother
\section{Piezoelectric Candidates}
\makeatletter\let\@fb@secFB\orbitol@secFB\makeatother
\label{sec:exp_piezo}

We apply skeleton-conditioned generation to the search for piezoelectric
candidates and evaluate the selected structures with density functional
theory (DFT). Piezoelectricity is allowed only in the 20 noncentrosymmetric crystal
classes other than 432 \citep{Aroyo2016InternationalTables}, which a generator
that accepts a space group can target directly.

We generated candidates with Orbitol, sampling skeletons from these
classes, relaxed the samples with NequIP-OAM-L as in
Section~\ref{sec:exp_setup}, and retained structures whose space group
survived relaxation and whose NequIP-OAM-L energy lay within 0.1~eV/atom
of the Alex-MP-20 DFT hull. A machine-learned screening pipeline, which is beyond the
scope of this work and will be reported in a separate publication,
predicted the piezoelectric figure of merit of these structures. We
selected 62 structures with high predicted figure of merit, and evaluated
them with DFT. Four had PBE band gaps of at most 0.03~eV and were excluded
from the response calculations, which left 58 candidates in 26 space groups:
26 nitrides and oxynitrides, 23 oxides, 8 halides and one sulfide.

We relaxed each candidate with PBE and computed its elastic, dielectric
and piezoelectric tensors, the latter two with density functional
perturbation theory (DFPT; Appendix~\ref{apdx:piezo}). The figure of merit
is the transverse energy-harvesting product
$d_{31}g_{31}=d_{31}^2/(\varepsilon_0\varepsilon^{T}_{33})$, with the
permittivity $\varepsilon^{T}$ at fixed stress
\citep{Bowen2014Harvesting}, maximized over orientations, which we also
evaluated on the 702 materials of the Materials Project's DFPT set with a
non-zero piezoelectric tensor
\citep{deJong2015Piezo,Petousis2017Dielectric,deJong2015Elastic}.

\begin{figure}[t]
    \centering
    \includegraphics[width=\linewidth]{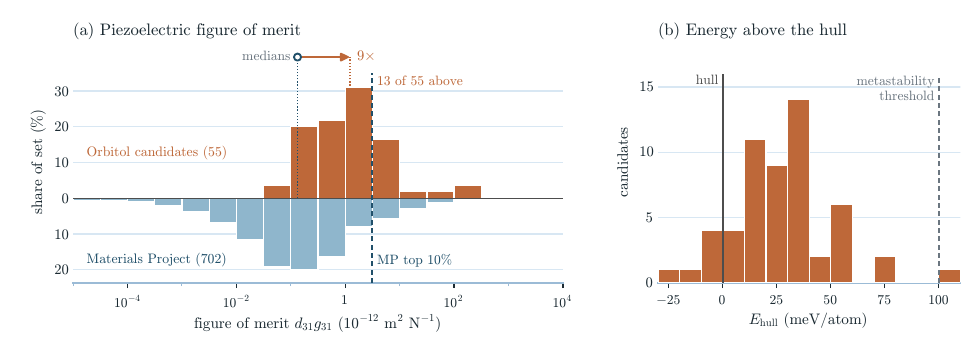}
    \caption{Piezoelectric candidates under DFT. (a) Figure of merit
    $d_{31}g_{31}$, maximized over orientations. Bars above the baseline show
    the 55 Orbitol candidates that pass both stability checks; bars below it
    show the 702 Materials Project materials with a non-zero piezoelectric
    tensor. Each bar covers a factor of $\sqrt{10}$ in $d_{31}g_{31}$, and its
    height is the percentage of the set in that range; three MP values below
    $10^{-5}$ are counted in the first bar. The dashed line marks MP's 90th
    percentile, 3.09; the arrow joins the two medians. (b) Energy above the
    hull of the MP and Alexandria phases for the same 55 candidates, in
    10~meV/atom bins. The dashed line marks the metastability threshold of
    0.1~eV/atom.}
    \label{fig:piezo}
\end{figure}

Because our DFT relaxations kept the crystal symmetry, we test its
stability with $\Gamma$-point phonons, which include symmetry-lowering
distortions (Appendix~\ref{apdx:piezo}). One candidate had an unstable mode
in its generated group $Cmc2_1$; following it lowered the symmetry to
$P2_1$. All 58 final structures are in piezoelectric crystal classes, and 56
have no imaginary optical mode at $\Gamma$. One of these 56 is mechanically
unstable; the remaining results refer to the 55 that pass both checks. Against the MP
and Alexandria phases
\citep{jainCommentaryMaterialsProject2013,Schmidt2023AlexandriaPBE},
recomputed with our settings, 54 of them lie within 0.1~eV/atom of the
hull, 46 within 50~meV/atom and 6 on or below it; the median is
27~meV/atom (Figure~\ref{fig:piezo}b). Of the 55, 49 do not match
any MP or Alexandria entry with the same space group.

The distribution of the figure of merit is shifted to larger values than
that of the MP set (Figure~\ref{fig:piezo}a): the median of the candidates
is $1.23\times10^{-12}$~m$^2$\,N$^{-1}$, against 0.133 for MP, and 13 of the
55 exceed MP's 90th percentile of 3.09. The shift reflects how the
candidates were chosen. The MP set was not selected for its response,
whereas our candidates were preselected for a high predicted figure of
merit, and only structures ranked high on it went on to DFPT.
Compared only with MP materials of the same point group, the candidates
have a median percentile of 72. Without the six that match Alexandria
structures, the median figure of merit of the other 49 is 1.23, and 12 of
them exceed MP's 90th percentile.

%% file: sections/conclusion.tex
\let\OrigFloatBarrier\FloatBarrier\let\FloatBarrier\relax
\section{Conclusion}
\label{sec:conclusion}
\let\FloatBarrier\OrigFloatBarrier

Orbitol generates crystals on Wyckoff orbits: given a skeleton, it generates
the atom type and free coordinates of each orbit and the free lattice
entries, and the decoder builds the requested symmetry into every valid
structure. On MP-20, it reached the highest MSUN and joint yield among the
evaluated symmetry-aware models, with symmetry fidelity after relaxation
comparable to the best baselines, while sampling orders of magnitude faster.
Generating directly in piezoelectric space groups, it produced candidates
that are both new and strongly responding: most of the screened structures
passed our DFT checks and match no MP or Alexandria entry with the same space
group, and their median figure of merit is about nine times that of the
Materials Project's DFPT piezoelectrics.

Fast generation opens directions that are expensive with current
symmetry-aware generators. In large-scale discovery campaigns, generation is
no longer the bottleneck, and reinforcement-learning fine-tuning, which needs
many fresh generation trajectories for every update, becomes affordable. Two
limits remain: the empirical prior samples only skeletons seen in training,
which learned skeleton priors could relax, and the benchmark energies are
MLIP estimates, while the piezoelectric candidates are DFT verified.

%% file: sections/appendix_classification.tex
\section{Classification of crystal generators}
\label{apdx:classification}

\begin{figure}[!h]
    \nocite{HadziVeljkovic2026Crystalite,yiCrystalDiTDiffusionTransformer2025a,xieCrystalDiffusionVariational,%
        jiaoCrystalStructurePrediction2024,millerFlowMMGeneratingMaterials2024,%
        zeniGenerativeModelInorganic2025,hoellmerOpenMaterialsGeneration2025,Seong2026MCFlow,%
        Okhotin2025MiAD,Edamadaka2026ALM,joshi2025allatom,Antunes2024CrystaLLM,%
        xu2025plaidpreferencealignedlanguage,Das2025TGDMat,Bone2025CrystaLLMpi,caoSpaceGroupInformed2025,%
        DeBreuck2026MatraGenoa,Chang2025SGEquiDiff,Lou2026GEODE,Cao2025CrystalFormerCSP,Zhou2026UFOMGen,%
        Karmush2026SymADiT,kazeevWyckoffTransformerGeneration2025,Zhu2024WyCryst,Kelvinius2025WyckoffDiff,%
        Jin2026DynaCrys,jiao2024space,Puny2025SGFM,Ruple2026SymmBFN,levy2025symmcd}%
    \centering
    \includegraphics[width=\linewidth]{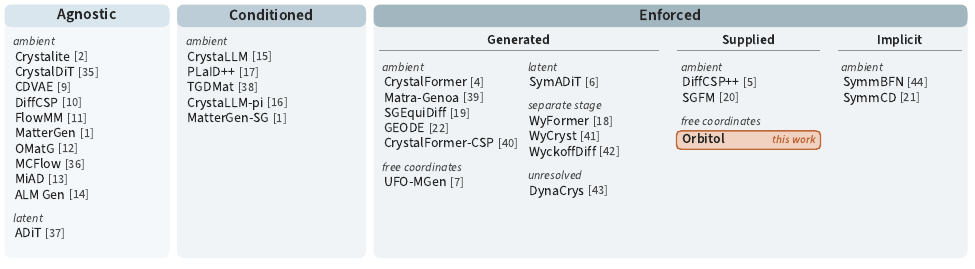}
    \caption{How crystal generators use the space group. Agnostic models do
    not specify it; Conditioned models receive or declare it without jointly
    constraining positions and lattice; Enforced models build it into the
    generated structure. Enforced models are divided by how they obtain
    the Wyckoff assignment (generated, supplied or implicit) and grouped by
    the coordinates they generate. Numbers in brackets are references.
    Table~\ref{tab:symmetry_models} lists the task, model family and lattice
    of each model.}
    \label{fig:taxonomy}
\end{figure}

This appendix classifies the crystal generators discussed in Section~\ref{sec:symmetry_models} by how
they use the space group. Figure~\ref{fig:taxonomy} groups them, and Table~\ref{tab:symmetry_models}
lists the task, model family, Wyckoff assignment, coordinates and lattice of each model.

\input{tables/taxonomy_models}

SymADiT's flow is an ADiT-based Transformer over learned orbit latents, and
its ambient-coordinate decoder ignores constrained slots
\citep{Karmush2026SymADiT}. GEODE's coordinate training target is a wrapped-normal
score restricted to reachable periodic images \citep{Lou2026GEODE}.

Three entries need qualification because their released code differs from
the paper.

UFO-MGen's classification follows its paper, which describes learned topology
and occupancy followed by flow in a masked free-coordinate chart
\citep{Zhou2026UFOMGen}.
The released command instead supplies the group and scaffold and uses
placeholder atom types. It zeroes the non-free coordinate slots after each
step, where the paper keeps them fixed by the mask, and, as in the paper,
projects the lattice only after sampling.
It establishes neither the full DNG procedure, whose composition source is
unspecified in the paper, nor enforcement for trigonal groups, where
rhombohedral lattice metrics are combined with hexagonal-axis Wyckoff
operations without conversion.

SymmCD's classification describes its intended projection onto compatible
Wyckoff positions; the release retains an unprojected representative if
projection raises an exception \citep{levy2025symmcd}.

Matra-Genoa's paper describes free Wyckoff parameters and rejection of
incompatible sequences \citep{DeBreuck2026MatraGenoa}.
We classify its released state as ambient with a family-constrained lattice:
it generates full coordinate triplets, resolves constrained components during
decoding, and projects the lattice to the crystal family before validity filtering.

%% file: tables/taxonomy_models.tex
\begin{table*}[!htbp]
\centering
\caption{Classification of the crystal generators discussed in this paper, grouped as in Figure~\ref{fig:taxonomy}. A dagger marks evaluated models, including symmetry-only comparisons; Task describes the model's scope: de novo generation (DNG), crystal structure prediction (CSP) or both. Wyckoff distinguishes generated row identifiers, supplied rows and implicit assignments within Enforced models. Coordinates gives the evolving state, not the network input, and Lattice whether the lattice is free, constrained to the crystal family, or produced in a separate stage.}
\label{tab:symmetry_models}
\small
\setlength{\tabcolsep}{6pt}
\begin{tabular}{@{}l lll ll@{}}
\toprule
Model & Task & Family & Wyckoff & Coordinates & Lattice \\
\midrule
\multicolumn{6}{@{}l}{\emph{Agnostic}} \\
ADiT \citep{joshi2025allatom} & DNG & Diffusion & n/a & Latent & Free \\
ALM Gen, placeholder prompt \citep{Edamadaka2026ALM}$^{\dagger}$ & DNG & Diffusion & n/a & Ambient & Free \\
CDVAE \citep{xieCrystalDiffusionVariational} & DNG & VAE & n/a & Ambient & Free \\
CrystalDiT \citep{yiCrystalDiTDiffusionTransformer2025a} & DNG & Diffusion & n/a & Ambient & Free \\
Crystalite \citep{HadziVeljkovic2026Crystalite}$^{\dagger}$ & Both & Diffusion & n/a & Ambient & Free \\
DiffCSP \citep{jiaoCrystalStructurePrediction2024}$^{\dagger}$ & Both & Diffusion & n/a & Ambient & Free \\
FlowMM \citep{millerFlowMMGeneratingMaterials2024} & Both & Flow & n/a & Ambient & Free \\
MatterGen (base) \citep{zeniGenerativeModelInorganic2025}$^{\dagger}$ & DNG & Diffusion & n/a & Ambient & Free \\
MCFlow \citep{Seong2026MCFlow} & Both & Flow & n/a & Ambient & Free \\
MiAD \citep{Okhotin2025MiAD}$^{\dagger}$ & DNG & Diffusion & n/a & Ambient & Free \\
OMatG \citep{hoellmerOpenMaterialsGeneration2025} & Both & Stoch.\ interp. & n/a & Ambient & Free \\
\addlinespace
\multicolumn{6}{@{}l}{\emph{Conditioned}} \\
CrystaLLM \citep{Antunes2024CrystaLLM}$^{\dagger}$ & Both & LLM & n/a & Ambient & Free \\
CrystaLLM-$\pi$ \citep{Bone2025CrystaLLMpi}$^{\dagger}$ & Both & LLM & n/a & Ambient & Free \\
MatterGen (SG adapter) \citep{zeniGenerativeModelInorganic2025} & DNG & Diffusion & n/a & Ambient & Free \\
PLaID++ \citep{xu2025plaidpreferencealignedlanguage}$^{\dagger}$ & DNG & LLM & n/a & Ambient & Free \\
TGDMat \citep{Das2025TGDMat} & Both & Diffusion & n/a & Ambient & Free \\
\addlinespace
\multicolumn{6}{@{}l}{\emph{Enforced}} \\
CrystalFormer \citep{caoSpaceGroupInformed2025}$^{\dagger}$ & DNG & Autoregressive & Generated & Ambient & Family \\
CrystalFormer-CSP \citep{Cao2025CrystalFormerCSP} & CSP & Autoregressive & Generated & Ambient & Family \\
DynaCrys \citep{Jin2026DynaCrys}$^{a}$ & DNG & Diffusion & Generated & Unresolved & Family \\
GEODE \citep{Lou2026GEODE} & DNG & Diffusion & Generated & Ambient & Family \\
Matra-Genoa \citep{DeBreuck2026MatraGenoa}$^{b}$ & DNG & Autoregressive & Generated & Ambient & Family \\
SGEquiDiff \citep{Chang2025SGEquiDiff}$^{\dagger,c}$ & Both & Diffusion & Generated & Ambient & Family \\
SymADiT \citep{Karmush2026SymADiT} & DNG & Flow & Generated & Latent & Family \\
UFO-MGen \citep{Zhou2026UFOMGen}$^{d}$ & DNG & Flow & Generated & Free coordinates & Family \\
WyckoffDiff \citep{Kelvinius2025WyckoffDiff} & DNG & Diffusion & Generated & Separate stage & Separate stage \\
WyCryst \citep{Zhu2024WyCryst} & Both & VAE & Generated & Separate stage & Separate stage \\
WyFormer \citep{kazeevWyckoffTransformerGeneration2025}$^{\dagger,e}$ & DNG & Autoregressive & Generated & Separate stage & Separate stage \\
DiffCSP++ \citep{jiao2024space}$^{\dagger}$ & Both & Diffusion & Supplied & Ambient & Family \\
SGFM \citep{Puny2025SGFM}$^{\dagger}$ & Both & Flow & Supplied & Ambient & Family \\
SymmBFN \citep{Ruple2026SymmBFN}$^{f}$ & Both & BFN & Implicit & Ambient & Family \\
SymmCD \citep{levy2025symmcd}$^{\dagger,g}$ & DNG & Diffusion & Implicit & Ambient & Family \\
Orbitol (ours)$^{\dagger,h}$ & DNG & Diffusion & Supplied & Free coordinates & Family \\
\bottomrule
\end{tabular}
\par\smallskip
\begin{minipage}{\textwidth}
\footnotesize
Evaluated Conditioned text models declare a group and expand positions with its operations, but leave the lattice unconstrained. Wyckoff applies only to Enforced models; n/a does not imply absence of Wyckoff text.
$^{a}$The paper generates the free coordinates of each orbit anchor with a DiffCSP++-style denoiser that acts on the full cell and projects noise and updates onto the free directions; it does not say whether the evolving state is the anchor position or its free coordinates.
$^{b}$Released implementation; state qualification follows the table.
$^{c}$Its CSP results use SGEquiDiff+LDiff with retrieved templates.
$^{d}$Paper classification; DNG follows its reported evaluations (no CSP benchmark). Release qualification follows the table.
$^{e}$The evaluated pipeline uses DiffCSP++ for geometry and lattice \citep{jiao2024space}.
$^{f}$For CSP, the space group and the atom types in the asymmetric unit are given.
$^{g}$Intended projection algorithm; release qualification follows the table.
$^{h}$Wyckoff assignments come from training skeletons. Both variants evolve the free coordinates; Orbitol-X takes ambient representative positions as network input and Orbitol-I the free coordinates themselves (Appendix~\ref{apdx:implementation_inputs}).
\end{minipage}
\end{table*}

%% file: sections/appendix_implementation.tex
\section{Orbitol definitions and implementation}
\label{apdx:implementation}

This appendix gives the definitions and implementation choices behind
Section~\ref{sec:method}: a detailed example of Wyckoff rows, the Wyckoff tables and
lattice charts, the coordinate metric and noise, the network inputs and conditioning, the
training target and losses, geometry in orbit attention, sampling and decoding, and the
reduction to Crystalite in P1.

\subsection{Detailed example of Wyckoff rows}
\label{apdx:wyckoff_example}

\begin{table}[!htb]
\centering
\small
\caption{The Wyckoff rows of $P4_2/mnm$ (No.~136) in the standard setting and the
orbits that rutile TiO$_2$ (mp-2657) and CrF$_4$ (mp-20488), both from the MP-20
training split, place on them. Free coordinates are given in the table convention,
$x=u+\tfrac12$, rounded to three decimals.}
\label{tab:wyckoff_example}
\begin{tabular}{@{}lccllll@{}}
\toprule
Row $w$ & $m_w$ & $d_w$ & Site symmetry & Representative & Rutile TiO$_2$ & CrF$_4$ \\
\midrule
$2a$ & 2 & 0 & $m.mm$ & $(0,0,0)$ & Ti & \\
$2b$ & 2 & 0 & $m.mm$ & $(0,0,\tfrac12)$ & & \\
$4c$ & 4 & 0 & $2/m..$ & $(0,\tfrac12,0)$ & & \\
$4d$ & 4 & 0 & $\bar 4..$ & $(0,\tfrac12,\tfrac14)$ & & \\
$4e$ & 4 & 1 & $2.mm$ & $(0,0,z)$ & & \\
$4f$ & 4 & 1 & $m.2m$ & $(x,x,0)$ & O, $x=0.305$ & Cr, $x=0.130$ \\
$4g$ & 4 & 1 & $m.2m$ & $(x,\bar x,0)$ & & F, $x=0.096$; F, $x=0.376$ \\
$8h$ & 8 & 1 & $2..$ & $(0,\tfrac12,z)$ & & \\
$8i$ & 8 & 2 & $m..$ & $(x,y,0)$ & & F, $(x,y)=(0.106,0.330)$ \\
$8j$ & 8 & 2 & $..m$ & $(x,x,z)$ & & \\
$16k$ & 16 & 3 & $1$ & $(x,y,z)$ & & \\
\midrule
\multicolumn{5}{@{}l}{Skeleton} & $\{2a,4f\}$ & $\{4f,4g,4g,8i\}$ \\
\multicolumn{5}{@{}l}{Atoms in the cell} & $2+4=6$ & $4+4+4+8=20$ \\
\bottomrule
\end{tabular}
\end{table}

Table~\ref{tab:wyckoff_example} lists every Wyckoff row of $P4_2/mnm$, the group of
rutile in Figure~\ref{fig:orbit_concepts}. The group fixes these eleven rows; a skeleton
selects a few of them, possibly more than once. Rutile uses two rows, $2a$ and $4f$.
CrF$_4$ uses three rows for four orbits: two F orbits lie on row $4g$ at different $x$.
Their tokens share the chart $(J_{4g},b_{4g})$ and the member maps of
Equation~\eqref{eq:orbit_members}, but each has its own atom type and free
coordinate, so moving one orbit leaves the other in place. A fixed row such as $2a$
cannot be occupied by a second orbit: with $d_{2a}=0$, it would occupy the same two points. The
four orbit tokens of CrF$_4$ have five free coordinates in total, and its lattice token
specifies the tetragonal lengths $a$ and $c$.

\subsection{Wyckoff tables, preprocessing and lattice charts}
\label{apdx:implementation_tables}
\label{apdx:implementation_lattice}

The Wyckoff tables use \lib{PyXtal}'s default Hall setting for each space group.
Row identifiers, charts, member operations, and lattice families use that
setting; trigonal groups use hexagonal axes.
For each row $w$, the tables give its conventional multiplicity $m_w$,
free-coordinate dimension $d_w$, member maps $(W_{w,j},v_{w,j})$ for
$j=1,\ldots,m_w$, and the representative chart $(J_w,b_w)=(W_{w,1},v_{w,1})$,
the map of the first member.
Chart tensors have three slots; only the first $d_w$ are active.
Padding and inactive coordinates are masked separately, and fixed rows
receive neither coordinate noise nor coordinate loss.

Preprocessing standardizes the cell, identifies its Wyckoff orbits, and
checks the atom types and geometry after expanding the tokens. Only structures
that pass this round-trip check enter the orbit cache.
We write $\mathrm{mod}_1(x)=x-\lfloor x\rfloor$ and denote the conventional
fractional members $f_{kj}$ of Equation~\eqref{eq:orbit_members} by
$f^{\mathrm{conv}}_{kj}$. The tables describe conventional cells. In a
centered lattice (I, C, F or R), the conventional cell contains $c_G>1$
lattice points, and members that differ by a centering translation are
periodic images of the same atom of the primitive cell. Primitive-member flags
keep one of them, so an orbit has $m_w^{\mathrm p}=m_w/c_G$ primitive members
\citep{Aroyo2016InternationalTables}.
Let $P_c$ contain the primitive basis vectors as columns in conventional
fractional coordinates. With column fractional positions and row lattice
vectors, the conventional lattice $L_{\mathrm{conv}}$ and positions convert
to the primitive lattice $L_{\mathrm p}$ and positions $f^{\mathrm p}_{kj}$ as
\begin{equation}
 L_{\mathrm p}=P_c^\top L_{\mathrm{conv}},\qquad
 f^{\mathrm p}_{kj}=\mathrm{mod}_1(P_c^{-1}f^{\mathrm{conv}}_{kj}).
 \label{eq:orbit_primitive}
\end{equation}
The output atom count is $N_{\mathrm p}=\sum_k m_{w_k}^{\mathrm p}$.

Augmentation composes proper normalizer maps, allowed polar-origin shifts,
and equivalent chart images. P1 instead receives a uniform global
translation. These operations change the description of the same crystal.

We use Crystalite's lower-triangular lattice latent $y$, reconstructed from
free slots $z$ by the crystal-family chart $y=\ell_G(z)$:
\begin{equation}
 L_{\mathrm{conv}}(z)=
 \begin{bmatrix}
 e^{y_1}&0&0\\
 y_2&e^{y_3}&0\\
 y_4&y_5&e^{y_6}
 \end{bmatrix}.
 \label{eq:orbit_lattice}
\end{equation}
Triclinic cells retain every slot. Monoclinic cells in the unique-axis-$b$
setting fix $y_2=y_5=0$; orthorhombic cells also fix $y_4=0$.
Tetragonal cells also tie $y_3=y_1$, and cubic cells tie $y_6=y_1$.
Hexagonal and trigonal groups use hexagonal axes, with
$y_2=-e^{y_1}/2$, $y_3=y_1+\log(\sqrt{3}/2)$, and $y_4=y_5=0$.

The free lattice state uses six padded slots. The network embeds the
reconstructed noisy lattice latent after EDM input scaling and predicts six
values. The fixed readout $\Gamma_G$ averages the first and third predictions
for tetragonal and hexagonal families, and also the sixth for cubic cells;
it then masks constrained slots. The lattice loss weights the shared diagonal
by the number of diagonal entries it represents. The dependent hexagonal
shear has no separate loss. State, noise, and readout use the same free-slot mask.

For the hexagonal chart, the shared log length is clamped to
$[-5-\log(\sqrt{3}/2),\log 20]$ before rebuilding the dependent entries.
At the sampling endpoint, free log-diagonal slots are clamped
to $[-5,5]$ and free shear slots to $[-10,10]$. Reconstruction then applies
the family chart, including its hexagonal bound.

\subsection{Chart metric, noise and preconditioning}
\label{apdx:chart_metric}
\label{apdx:implementation_noise}

Fractional rotations need not preserve Euclidean lengths. We therefore use
one group-invariant metric, fixed independently of the noisy lattice, to
measure displacements and set chart noise. For point-group rotations
$\mathcal R_G$, define
\begin{align}
 M_G&=\frac{1}{|\mathcal R_G|}\sum_{R\in\mathcal R_G}R^\top R,
 &H_w&=J_w^\top M_G J_w,\\
 \Pi_w&=H_w^{-1}J_w^\top M_G,
 &N_w&=H_w^{-1}J_w^\top M_G^{1/2}.
 \label{eq:orbit_metric}
\end{align}
Right multiplication by a point-group rotation $Q$ permutes the group, so
\begin{equation*}
 Q^\top M_GQ=\frac{1}{|\mathcal R_G|}
 \sum_{R\in\mathcal R_G}(RQ)^\top(RQ)=M_G.
\end{equation*}
Thus a chart step $\delta$ gives every member the same squared displacement
$\|R_jJ_w\delta\|_{M_G}^2=\delta^\top H_w\delta$ before wrapping.
The coordinate noise of Equation~\eqref{eq:orbit_noising} is
$\xi_k=N_{w_k}\epsilon_{u,k}$ with a standard Gaussian
$\epsilon_{u,k}\in\mathbb R^3$. On active coordinates, $N_wN_w^\top=H_w^{-1}$,
so $\sigma N_w\epsilon$ has covariance $\sigma^2H_w^{-1}$, or $\sigma^2I$ after multiplication by
$H_w^{1/2}$. Fixed rows receive no coordinate noise.
Minimizing $\|J_w\delta-F\|_{M_G}^2$ gives
$H_w\delta=J_w^\top M_GF$, hence the readout $\delta=\Pi_wF$;
for $J_w=I$, $H_w=M_G$ and $\Pi_w=I$.

For example, rutile has $M_G=I$ and $H_{4f}=2$: a chart step $\delta u$ on
row 4f moves each of the four O members by an $M_G$ length
$\sqrt{2}|\delta u|$ before wrapping.
For a nonorthogonal example, point group $6/mmm$ in hexagonal axes has the
24 operations $R^kS^sZ^z$, $k=0,\ldots,5$, $s,z\in\{0,1\}$, where
\begin{equation*}
 R=\begin{pmatrix}1&-1&0\\1&0&0\\0&0&1\end{pmatrix},\quad
 S=\begin{pmatrix}0&1&0\\1&0&0\\0&0&1\end{pmatrix},\quad
 Z=\operatorname{diag}(1,1,-1).
\end{equation*}
Their average gives
$M_G=\left(\begin{smallmatrix}4/3&-2/3&0\\-2/3&4/3&0\\0&0&1\end{smallmatrix}\right)$.
The rotation $R$ takes $(1,0,0)^\top$ to $(1,1,0)^\top$, changing its
Euclidean length from $1$ to $\sqrt2$ but preserving its squared $M_G$
norm of $4/3$.
The identity metric would therefore assign different displacement lengths
to members of the same orbit.

We use Crystalite's EDM formulation
\citep{NEURIPS2022_a98846e9,HadziVeljkovic2026Crystalite}, drawing one shared
noise level per structure from a clipped log-normal distribution and
independent Gaussian draws for the channels in
Equation~\eqref{eq:orbit_noising}, with free-slot masks.
Write $\widetilde u_k^\sigma=u_k+\sigma\xi_k$ for the
unwrapped coordinate state: objective M retains it during training,
whereas target P uses $u_k^\sigma=\widetilde u_k^\sigma-
\mathrm{round}(\widetilde u_k^\sigma)$.
For channel $a\in\{h,u,z\}$ and data scale $\sigma_{\mathrm{data},a}$,
\begin{equation}
 \begin{aligned}
 c_{\mathrm{skip},a}&=\frac{\sigma_{\mathrm{data},a}^2}{\sigma^2+\sigma_{\mathrm{data},a}^2},&
 c_{\mathrm{out},a}&=\frac{\sigma\sigma_{\mathrm{data},a}}{\sqrt{\sigma^2+\sigma_{\mathrm{data},a}^2}},\\
 c_{\mathrm{in},a}&=\frac{1}{\sqrt{\sigma^2+\sigma_{\mathrm{data},a}^2}}.
 \end{aligned}
 \label{eq:orbit_edm_coefficients}
\end{equation}
These coefficients give the type and lattice denoisers in
Equation~\eqref{eq:orbit_denoisers}; coordinates receive no input scaling.
The coordinate target and skip are separate choices, specified below.

\subsection{Network inputs, conditioning and coordinate variant}
\label{apdx:implementation_inputs}
\label{apdx:implementation_conditioning}

\emph{Orbitol-X} is the model called Orbitol in the main text and used for
all its MP-20 results. \emph{Orbitol-I} changes the coordinate input and
readout; Table~\ref{tab:ablations} compares the variants.
Orbitol-X applies Crystalite's Fourier embedder to the representative
$f_{k1}$ in Equation~\eqref{eq:orbit_members}, evaluated at $u_k^\sigma$ for
P or $\widetilde u_k^\sigma$ for M. Its coordinate readout is
$\widehat F_{u,k}=\Pi_{w_k}F_{u,k}$; the ambient input adds no free variables.

Orbitol-I instead embeds $\mathrm{mod}_1(u_k^\sigma+\tfrac12)$ with Fourier
features, masking inactive-coordinate features before the embedding MLP.
Fixed rows retain Orbitol-X's Fourier features of $\mathrm{mod}_1(b_w)$.
This input is periodic under integer chart lifts, but does not identify
every additional fractional translation in $\Lambda_w$.
Its direct readout is $\widehat F_{u,k}=H_{w_k}^{-1/2}F_{u,k}$, with inactive
output slots masked.

The token and group embeddings follow Equations~\eqref{eq:token_inputs}
and~\eqref{eq:group_conditioning}, with the lattice token last and padding
masked. Both variants use hybrid row and group embeddings, each the sum of
a learned lookup and an MLP on standardized physical descriptors.
Row descriptors include log multiplicity, free-coordinate dimension,
reciprocal stabilizer size, operation counts, and entries of
$J_wH_w^{-1}J_w^\top$. Group descriptors encode crystal system, centering,
point-group operations, screw and glide information, Wyckoff statistics,
and spectral coordinates of subgroup relations.
The group embedding is added to the noise embedding of
$c_{\mathrm{noise}}=\tfrac14\log\sigma$ in every AdaLN block
\citep{peeblesScalableDiffusionModels2023}.
Categorical additions and final descriptor-MLP layers start at zero;
the charts enforce the constraints even when these embeddings vanish.

\subsection{Periodic target and training losses}
\label{apdx:implementation_objectives}

The benchmark models and Orbitol-X/I comparison use the periodic-posterior target P of
Equation~\eqref{eq:orbit_periodic_target}, with the sine skip
\begin{equation}
 \begin{aligned}
 s_u(u)&=u-(1-c_{\mathrm{skip},u})\frac{\sin(2\pi u)}{2\pi},\\
 D_{u,k}^{P}&=s_u(u_k^\sigma)+c_{\mathrm{out},u}\widehat F_{u,k}.
 \end{aligned}
 \label{eq:orbit_periodic_denoiser}
\end{equation}
The sine acts componentwise, giving $s_u(u+n)=s_u(u)+n$ for integer $n$
and matching the EDM skip to first order near the chart origin.
The skip comparison also uses P with $s_u(u)=c_{\mathrm{skip},u}u$.
Table~\ref{tab:experiment_configs} lists which models use each skip.
Writing $p_\sigma$ for the wrapped transition density, the target satisfies
\begin{equation*}
 D^\star=u^\sigma+\sigma^2H_w^{-1}
 \nabla_{u^\sigma}\log p_\sigma(u^\sigma\mid u).
\end{equation*}
Regression averages this conditional lifted mean over clean points
compatible with the noisy channels. At high noise the wrapped density
becomes uniform and $D^\star$ approaches $u^\sigma$, without preferring a
chart origin. The target sums integer lifts only; augmentation samples the other
equivalent chart images, including fractional pure translations.
The clean point is used to construct the training target, while the
denoiser receives only noisy channels and the fixed skeleton.
The coordinate loss is
\begin{equation}
 \mathcal L_u=
 \left\langle c_{\mathrm{out},u}^{-2}
 \|D_{u,k}^{P}-D^\star(u_k^\sigma,u_k)\|_{H_{w_k}}^2
 \right\rangle_{\mathrm{active\ orbits}},
 \label{eq:orbit_coordinate_loss}
\end{equation}
without wrapping the prediction error.

For comparison, objective M uses
$D_{u,k}^{M}=c_{\mathrm{skip},u}\widetilde u_k^\sigma+
 c_{\mathrm{out},u}\widehat F_{u,k}$ and selects the nearest residual over
$\Lambda_w$, the lattice generated by integer periods and translation-only
chart images:
\begin{equation}
 \mathrm{wrap}_w(\delta)=\delta-\operatorname*{arg\,min}_{\lambda\in\Lambda_w}
 \|\delta-\lambda\|_{H_w}^2.
 \label{eq:orbit_wrapped_residual}
\end{equation}
Its per-orbit error is
$c_{\mathrm{out},u}^{-2}\|\mathrm{wrap}_w(D_u^M-u)\|_{H_w}^2$.
M selects the nearest periodic image, whereas P averages the periodic images with
their conditional Gaussian weights. Their conditional optima need not
agree, so M need not recover the posterior mean used by the diffusion drift.
This is an analytic distinction; we do not compare trained M and P models.
Periodic-score precedents are discussed in Section~\ref{sec:method_targets}.

Both coordinate losses average over real orbit tokens with $d_w>0$ across
the batch, without dividing by coordinate dimension or weighting by
multiplicity; an empty active set contributes zero.
Type error is summed over the 16 type features and averaged over real
orbit tokens. Lattice error uses the tied-diagonal weights of
Appendix~\ref{apdx:implementation_lattice} and is averaged over all six
slots and structures. Each channel receives its EDM weight
$c_{\mathrm{out},a}^{-2}$; the channel weights in the total loss are given
in Appendix~\ref{apdx:experiments}.

The infinite sum in Equation~\eqref{eq:orbit_periodic_target} is evaluated
in double precision with adaptive image sums at low noise, a dual-lattice
Fourier series at higher noise, and an identity limit when the correction
is negligible. Training caches metric-dependent primal and dual grids and
masks terms by noise level, without a universal fixed image-count cutoff.
The batched implementation uses padded metric tables for all tokens and
agrees with the loop implementation within $10^{-12}$ absolute tolerance
in double precision.

\subsection{Geometry in orbit attention}
\label{apdx:implementation_gem}
\label{apdx:gem_intuition}

We expand the noisy orbits into their primitive members $a$, with fractional
coordinates $f_a$ and owner orbit $o(a)$; $r(k)$ is the first member of query
orbit $k$.
For distance evaluation, let $L_s$ be the sanitized primitive cell and
$L_r=TL_s$ its reduced basis, with integer unimodular $T$.
Before computing distances, we replace nonfinite primitive cells with the
identity cell. Finite row lengths are bounded to
$[e^{-5},e^5]\,\text{\AA}$ while preserving directions; a degenerate
row-normalized cell also falls back to the identity. Nonfinite fractional
coordinates become zero, and coordinates are wrapped.
Sorted size reduction produces $L_r=TL_s$ with a bounded iteration count
and an identity-transform fallback. We compute the fractional displacements
$\delta_{ab}$, their wrapped versions $\bar\delta_{ab}$, and distances $d_{ab}$ as
\begin{align}
 \delta_{ab}&=T^{-\top}(f_b-f_a),
 &\bar\delta_{ab}&=\delta_{ab}-\mathrm{round}(\delta_{ab}),\\
 d_{ab}&=\min_{n\in\{-1,0,1\}^3}
 \|(\bar\delta_{ab}+n)^\top L_r\|_2.
 \label{eq:orbit_gem_distance}
\end{align}
The search uses the neighboring offsets in this reduced basis. It is a finite
nearest-image approximation, without a closest-vector guarantee for arbitrary
noisy cells.

For attention head $g$, using radial basis features $\mathrm{RBF}$ and volume
scale $\bar a=|\det L_s|^{1/3}$, the distance-only atom bias is
\begin{equation}
 \beta_g(a,b)=\mathrm{MLP}_g\!\left(
 [\mathrm{RBF}(d_{ab}/\bar a),\mathrm{RBF}(d_{ab})]\right),
 \label{eq:orbit_gem_member_bias}
\end{equation}
with $\beta_g(a,a)=0$.
Writing $d$ for a member-pair distance $d_{ab}$, the radial features are
Gaussian functions with evenly spaced centers,
using their spacing as width. The default uses 32 centers over $[0,2]$ for
$d/\bar a$ and 32 over $[0,8\,\text{\AA}]$ for $d$. We concatenate these
features and pass them to an MLP with a hidden width of 128, SiLU activation,
and one output per attention head. Its final layer starts at zero. Among real pairs, only the
identical-atom pair is explicitly assigned zero bias; padding is masked.
The scale $\bar a$ comes from the same sanitized cell as the distances.

Equation~\eqref{eq:orbit_gem_aggregation} sums key-member weights for one
query representative $r(k)$. The lattice cases are
$B_g[\mathrm{lat},l]=\log m_{w_l}^{\mathrm p}$ and
$B_g[k,\mathrm{lat}]=B_g[\mathrm{lat},\mathrm{lat}]=0$.
For members of orbit $l$ with tied key $\kappa_l$ and value $\nu_l$, set
$\alpha_{kl}=q_k^\top\kappa_l/\sqrt{d_h}$, with head width $d_h$ and
attention normalizer $Z_k$. Then
\begin{equation}
 \sum_{b:o(b)=l}\frac{e^{\alpha_{kl}+\beta_g(r(k),b)}}{Z_k}
 =\frac{e^{\alpha_{kl}+B_g[k,l]}}{Z_k}.
 \label{eq:orbit_tied_attention}
\end{equation}
This \emph{tied-attention identity} preserves the weighted value sum up to
rounding for any supplied atom biases; averaging key-member weights would
lose their total mass. Zero atom biases give the multiplicity-only case
$B_g[k,l]=\log m_{w_l}^{\mathrm p}$, including at MLP initialization.
With exact distances in a symmetry-compatible cell, changing the query
representative permutes each key orbit's distance multiset, but sanitization
and finite image searches need not preserve this invariance for arbitrary
noisy cells.

In rutile, Ti occupies row 2a and O row 4f of $P4_2/mnm$: GEM expands six
primitive atoms, while the Transformer processes two orbit tokens and a
lattice token. The following multipliers are illustrative distance-to-weight
values, not fitted MLP outputs; equal distances receive equal values and
self-pairs receive one. Rows are query atoms and columns are key atoms.
\begin{center}
\small
\begin{tabular}{@{}l|rr|rrrr@{}}
$e^{\beta_g(a,b)}$ & Ti$_1$ & Ti$_2$ & O$_1$ & O$_2$ & O$_3$ & O$_4$ \\
\hline
Ti$_1$ & 1 & 2 & 6 & 6 & 7 & 7 \\
Ti$_2$ & 2 & 1 & 7 & 7 & 6 & 6 \\
\hline
O$_1$ & 6 & 7 & 1 & 4 & 3 & 3 \\
O$_2$ & 6 & 7 & 4 & 1 & 3 & 3 \\
O$_3$ & 7 & 6 & 3 & 3 & 1 & 4 \\
O$_4$ & 7 & 6 & 3 & 3 & 4 & 1 \\
\end{tabular}
\qquad
\begin{tabular}{@{}l|rrr@{}}
$e^{B_g[k,l]}$ & Ti & O & lattice \\
\hline
Ti & 3 & 26 & 1 \\
O & 13 & 11 & 1 \\
lattice & 2 & 4 & 1 \\
\end{tabular}
\end{center}
Summing key columns for representatives Ti$_1$ and O$_1$ gives the reduced
table: the Ti-to-O multiplier is $6+6+7+7=26$, and changing representatives
only permutes these terms. The lattice query sums unit multipliers, giving
the primitive counts $(2,4)$, while its single key has multiplier one.
With content multipliers $(2,1,1)$ for a Ti query, expanded weights
$(2,4,6,6,7,7,1)$ and orbit weights $(6,26,1)$ both total $33$ and give
\begin{equation*}
 (p_{\mathrm{Ti}\to\mathrm{Ti}},p_{\mathrm{Ti}\to\mathrm O},
 p_{\mathrm{Ti}\to\mathrm{lat}})
 =\left(\frac{6}{33},\frac{26}{33},\frac{1}{33}\right).
\end{equation*}
Tied values therefore give the same attention output up to rounding.
This verifies aggregation for the supplied biases, independently of the
finite-distance approximation above.

Geometry is computed without gradients from the unscaled noisy state,
outside mixed precision; the bias MLP trains, and its output is shared
across layers with no explicit noise gate or noise-level input.
When compiled geometry is enabled, reduction uses eight sweeps and leaves
cells unreduced if more are needed, matching the eager reduction when it
converges within that bound.

Crystalite's GEM combines a distance penalty with Fourier displacement,
normalized-distance, and lattice features, then applies a noise-dependent
gate~\citep{HadziVeljkovic2026Crystalite}. It normalizes by mean cell length
and leaves lattice-token interactions unbiased. Orbitol uses absolute and
volume-normalized distances, sums key-member weights, and includes
multiplicity in the lattice-query row; its P1 limit therefore differs
from Crystalite's configured GEM.

\subsection{Sampling, decoding and valid outputs}
\label{apdx:implementation_sampling}

The sampler initializes Gaussian states with the training masks and chart
noise maps, then applies Crystalite's EDM schedule, stochastic churn and
Heun correction. P uses coordinate drift $(u^\sigma-D_u)/\sigma$; M wraps
the numerator over $\Lambda_w$. Coordinate states are wrapped componentwise
after Euler and Heun updates. With P and the EDM skip, denoising, drift
and updates use the same wrapped state, including after churn; with the
sine skip, integer-lift equivariance permits lifted churn states until the
update is wrapped. Channel-wise anti-annealing is supported but disabled
in all reported runs.

Sampler settings and the sensitivity study are given in
Appendix~\ref{apdx:sampler_sweep}.

At the endpoint, free lattice slots are clamped and dependent entries
rebuilt. Crystalite's type encoding decodes predicted features to allowed
types. Equations~\eqref{eq:orbit_members} and~\eqref{eq:orbit_primitive},
with primitive-member flags, produce the primitive structure without a
geometric primitive-cell search. The decoder checks that requested group
operations preserve the lattice and map atoms bijectively to same-type
periodic images. For evaluation, the primitive cell is Niggli-reduced,
falling back to LLL and then the input cell if reduction fails; evaluator
rows record the resulting metric under the final lattice clamp.

Active-state finiteness is recorded before endpoint sanitization; a false
flag causes rejection. The decoder also rejects invalid groups, rows or
types, incompatible token shapes, empty structures, nonfinite values,
invalid lattices, reused fixed rows, coinciding distinct orbits, collapsed
members within an orbit, and failed group containment. Evaluator conversion
records conversion failures and atom-capacity violations. Failures remain
in the output metadata and are not replaced when scoring attempts.

\subsection{P1 reduction and verification}
\label{apdx:implementation_parity}

In P1, every orbit is 1a with $m_w=1$, $d_w=3$,
$J_w=H_w=N_w=\Pi_w=I$ and $b_w=0$. Primitive and conventional cells
coincide, and the triclinic lattice chart is the identity.
Orbitol computes Crystalite's denoiser, loss and sampler exactly when type
encodings, weights, execution modes and sampler settings match, with zero
row and group embeddings, objective M and GEM off. Both evaluated P skips
depart from this equality even in P1; configured GEM differences are given
in Appendix~\ref{apdx:implementation_gem}.

Tests compared cached coordinates and lattice states, initialized model
outputs across embedding modes, and objective-M corruption, denoising and
losses at zero tolerance; decoded-cell round trips allowed numerical
conversion tolerance. Sampler tests used matched random streams with and
without anti-annealing and compared final type, coordinate and lattice
states at zero tolerance, without recording intermediate trajectories.
Separate integration tests trained and sampled under the implemented
coordinate objectives.

%% file: sections/appendix_experiments.tex
\section{Experimental details and supporting results}
\label{apdx:experiments}

This appendix supports Section~\ref{sec:experiments}: the evaluated configurations, how
yield and symmetry are scored, the SPARC derivation, the full results and reported
comparisons, the coordinate-configuration comparison, the sampler settings and the
timing of sampling.

\subsection{Data, training and evaluated configurations}

The MP-20 orbit training cache contains 27,136 structures. The input
Alex-MP-20 splits contain 540,162 training structures and 67,521 each for
validation and test; the Alex orbit training cache retains 540,159 structures.
We sample skeletons only from the training cache; each supplies the space
group and an untyped multiset
of Wyckoff rows. Orbitol does not copy the template's atom types or geometry.
The empirical skeleton prior does not generate unseen skeletons.

Table~\ref{tab:hyperparameters} lists the settings of the two benchmark models,
read from their checkpoint configurations. Training batches are padded rather than packed.
The main MP-20 run used compiled training steps and compiled GEM geometry,
batched periodic targets, device-side type encoding and fused AdamW/EMA updates.
The AdamW kernel reproduces \lib{PyTorch}'s foreach arithmetic with
single-precision parameters and moments, and the data assembly and type
encoding preserve the batches bitwise.
Table~\ref{tab:experiment_configs} distinguishes the evaluated configurations;
Appendices~\ref{apdx:implementation_inputs} and~\ref{apdx:implementation_objectives}
define their inputs, readouts and periodic-posterior target P.

\begin{table}[!htb]
\centering
\small
\caption{Settings of the two benchmark models (Orbitol-X). A value centered
across both columns is shared. Table~\ref{tab:experiment_configs} lists the
controls and the Orbitol-I variant; Appendix~\ref{apdx:evaluation} gives the
evaluation details.}
\label{tab:hyperparameters}
\begin{tabular}{@{}lcc@{}}
\toprule
Setting & MP-20 & Alex-MP-20 \\
\midrule
\multicolumn{3}{@{}l}{\emph{Architecture}} \\
Width, layers, heads, MLP ratio & \multicolumn{2}{c}{512, 14, 16, 4} \\
Parameters & \multicolumn{2}{c}{69.2M} \\
Maximum orbits per crystal & \multicolumn{2}{c}{20} \\
Coordinate input; readout & \multicolumn{2}{c}{representative, 32 Fourier frequencies; projection} \\
Group and row embeddings & \multicolumn{2}{c}{learned lookup vector and descriptor MLP} \\
GEM & \multicolumn{2}{c}{distance, shared across layers, no noise gate} \\
\addlinespace
\multicolumn{3}{@{}l}{\emph{Training}} \\
Target; coordinate skip & \multicolumn{2}{c}{P; sine} \\
Loss weights (type, coordinate, lattice) & (0.25, 50, 5) & (0.5, 50, 5) \\
EDM data scale; $\ln\sigma$ & \multicolumn{2}{c}{0.3 per channel; $\mathcal N(-1.2,\,1.2^2)$} \\
Optimizer & \multicolumn{2}{c}{AdamW, learning rate $10^{-4}$, no weight decay} \\
Batch size; learning-rate schedule & \multicolumn{2}{c}{128; 1,000 warm-up steps, cosine decay} \\
Gradient clipping; EMA decay & \multicolumn{2}{c}{1; 0.9999} \\
Precision & \multicolumn{2}{c}{bfloat16} \\
Training steps: schedule; evaluated & 2.5M; 1.5M & 1M; 650k \\
\addlinespace
\multicolumn{3}{@{}l}{\emph{Sampling}} \\
Skeleton prior & \multicolumn{2}{c}{empirical training skeletons, temperature 1} \\
Sampler; weights & \multicolumn{2}{c}{EDM stochastic Heun \citep{NEURIPS2022_a98846e9}; EMA} \\
Heun steps (network evaluations) & 25 (49) & 150 (299) \\
$[\sigma_{\min},\sigma_{\max}]$; $S_{\mathrm{churn}}$ & $[0.01,3]$; 10 & $[0.002,80]$; 60 \\
$\rho$, $S_{\min}$, $S_{\max}$, $S_{\mathrm{noise}}$ & \multicolumn{2}{c}{7, 0, 999, 1.003} \\
Execution & packed (8,192 tokens), compiled & padded batches of 256 \\
Precision; decoder & \multicolumn{2}{c}{bfloat16; reference} \\
Sampling seeds & 3 & 1 \\
\addlinespace
\multicolumn{3}{@{}l}{\emph{Evaluation}} \\
Attempts per seed & \multicolumn{2}{c}{2,500} \\
Relaxation & \multicolumn{2}{c}{NequIP-OAM-L, FIRE with Fr\'echet cell filter, 500 steps} \\
Symmetry detection & \multicolumn{2}{c}{\lib{spglib}, 0.1~\AA\ (also 0.01~\AA), angle $5^\circ$} \\
Scoring & \multicolumn{2}{c}{LeMat-GenBench, mean ORB, MACE and UMA $E_{\mathrm{hull}}$, raw fallback} \\
\bottomrule
\end{tabular}
\end{table}

\begin{table}[!htb]
\centering
\small
\caption{Evaluated Orbitol configurations. All use width 512, distance GEM,
target P and EMA weights. Updates are the evaluated checkpoint and full
training schedule; each sampling seed gives 2,500 attempts. The sampler column
gives Heun steps; full settings are in Table~\ref{tab:hyperparameters}.
Orbitol-X uses representative inputs and projection readout; Orbitol-I uses
chart-coordinate inputs and direct readout.}
\label{tab:experiment_configs}
\setlength{\tabcolsep}{4pt}
\begin{tabular}{@{}p{0.29\linewidth}lrrrr@{}}
\toprule
Dataset / role & Model / skip & $\lambda_{\rm type}$ & Updates & Sampler & Seeds \\
\midrule
MP-20 benchmark & X / sine & 0.25 & 1.5M / 2.5M & 25 & 3 \\
MP-20 control & X / sine & 1 & 400k / 500k & 150 & 1 \\
MP-20 control & X / EDM & 1 & 400k / 500k & 150 & 1 \\
Alex-MP-20 benchmark & X / sine & 0.5 & 650k / 1M & 150 & 1 \\
Alex-MP-20 application variant & I / sine & 0.5 & 650k / 1M & 150 & 1 \\
\bottomrule
\end{tabular}
\end{table}

We selected the main MP-20 checkpoint by the highest observed in-training
MSUN; for Alex-MP-20, we report the checkpoint at 650k steps. Each MP-20 screen drew 10,000 attempts, relaxed constructed structures for
500 FIRE steps, and measured uniqueness and novelty on the relaxed
endpoints against MP-20 train, with stability against the in-house MP-20 hull.
This selection score counts failures as zero and differs from the LeMat
benchmark. The MP-20 controls and the Alex-MP-20 application variant use the
checkpoints in Table~\ref{tab:experiment_configs}. Independent training replicates are unavailable.

\subsection{Evaluation and failure accounting}
\label{apdx:evaluation}

Table~\ref{tab:exp_metrics} summarizes the metrics of Section~\ref{sec:exp_protocol}.

\begin{table}[t]
\centering
\small
\caption{Evaluation metrics. All rates are percentages. Yield, requested
or declared retention, SPARC and joint yield use all
2,500 attempts. P1 and JSD use eligible relaxed detections; failures and
implausible endpoints are excluded from these distributions and count as
losses for retention, SPARC and joint yield.}
\label{tab:exp_metrics}
\begin{tabular}{@{}p{0.25\linewidth} p{0.70\linewidth}@{}}
\toprule
Metric & Definition \\
\midrule
SUN, MSUN & Attempts that are valid, unique among the model's 2,500 attempts and novel against
LeMat-Bulk, with predicted energy above hull at most 0 (SUN) or at most
0.1~eV/atom (MSUN, which adds the metastable candidates to SUN).
Yield counts all attempts with raw fallback. P1, JSD, retention and SPARC
use relaxed structures and are not restricted to those counted in MSUN. \\
P1 & Share of relaxed structures detected in P1, compared with the training
share (1.7\% on MP-20). \\
JSD & Jensen--Shannon divergence, in bits, between the space-group
distribution of the relaxed structures and that of the training set.
We compare detected distributions after relaxation; sampling training
skeletons sets the requested-group distribution, not the detected histogram. \\
Requested group retained & Attempts whose eligible relaxed structure is detected in the requested
group; for models that declare their own group, the declared group.
For retention and SPARC, relaxing into a supergroup counts as a loss. \\
Joint yield & Attempts that satisfy MSUN and retain the requested or declared
group after relaxation. Failures and implausible endpoints
count as not retained. \\
SPARC & Attempts that retain their raw group through relaxation, weighted by
how well the retained groups cover the training distribution
(Equation~\ref{eq:sparc}). \\
\bottomrule
\end{tabular}
\end{table}

We used LeMat-GenBench at its revision of 15 September 2026, which pins the
corrected MACE checkpoints, with its multi-MLIP hull configuration. ORB, MACE and UMA used their
respective hulls, and novelty was measured against LeMat-Bulk.
For SUN and MSUN, LeMat thresholds each structure's arithmetic mean of
the available ORB, MACE and UMA energies above their respective hulls at
0 and 0.1~eV/atom, respectively. Missing model values are omitted from
the mean; a structure without any value counts as neither stable nor
metastable.

For every evaluated model and seed, we selected attempts 0 through 2499 from the model's
pool, retaining generation, decoding and structure-check failures. We did
not replace failures or select a subset of parseable outputs. Yield and
requested-group retention use all 2,500 attempts as their denominator.
For PLaID++, each language-model generation is an attempt; generations that
fail to decode, which its released sampler discards and redraws, count as
failures (7.2\% of attempts). We used one sampling seed per model and three
for Orbitol-X (MP-20) and PLaID++, averaging each metric over the three seeds.
The main MP-20 table and Figure~\ref{fig:symmetry_distribution} both use
the train-only SymmCD and WyFormer retrainings. SGFM and Orbitol sample the same empirical untyped skeleton law,
although their output cell conventions and samplers differ.

We pre-relaxed structures with the \lib{OpenEquivariance} TorchScript build
of NequIP-OAM-L \citep{Kavanagh2026NequIPOAM}, using FIRE and a
Frechet cell filter for 500 fixed steps on one RTX 6000 Ada. Atomic and
cell updates had no symmetry constraint. LeMat used single-point energies
on these endpoints, with raw fallback when relaxation failed. Its
additional model-specific relaxations are separate outputs and did not
define the input geometry for the reported hull scores. Invalid attempts
contributed zero yield.

We used \lib{pymatgen}'s \texttt{SpacegroupAnalyzer} with \lib{spglib} 2.7.0, angle
tolerance $5^\circ$, and positional tolerances of 0.1~\AA\ and 0.01~\AA.
Shares and JSD use detected structures; detection failures are not P1.
Relaxed symmetry includes only successful endpoints and excludes
\emph{implausible endpoints}: a relaxed-to-raw volume ratio outside $[0.5,2]$, or an
interatomic distance below 0.5~\AA. These exclusions apply to relaxed
symmetry, separately from LeMat's raw fallback rule. We detected each
dataset's training reference with the same settings; the Alex-MP-20
reference contains 540,162 training structures. JSD uses base-two
logarithms and is reported in bits.

Requested-to-raw and requested-to-relaxed retention require exact equality
of the requested and detected groups, divided by all 2,500 attempts.
For CrystaLLM, CrystaLLM-$\pi$ and PLaID++, the request is replaced by the
model's own declaration. Generation, relaxation and detection failures,
and implausible relaxed endpoints, count as losses in the relevant rate.
The declaration is read from generated text, not a CIF serialization
header. Agnostic models have no requested-group retention.
\emph{Own-group strict retention} instead asks whether a structure retained
its raw-detected group after relaxation, using all raw-detected structures
as the denominator. Failed relaxations, implausible endpoints and failed relaxed
detections count as losses. \emph{Own-group joint retention} uses only
pairs with both raw and eligible relaxed detections. Neither denominator
is interchangeable with the attempt denominator for requested-group
retention, and none of these equality tests measures supergroup containment.
Both own-group rates use the same retained count as $R_{\rm raw}$, which
instead divides by all attempts (Appendix~\ref{apdx:sparc}).

We compute joint yield by mapping the
union of LeMat's SUN and disjoint metastable indices through the valid-input
IDs to sample IDs, then joining them to per-sample symmetry at 0.1~\AA;
failed relaxation or detection and implausible endpoints count as not retained.
We divide the joint count by all 2,500 attempts for joint yield and by the
MSUN count for retention among MSUN attempts, averaging each metric over
the seeds of Orbitol-X (MP-20) and PLaID++.

\subsection{SPARC}
\label{apdx:sparc}

Let $k_g$ be the number of attempts that retain group $g$, $K=\sum_g k_g$
the total retained count and $N$ the total number of attempts.
A retained attempt has the same detected group in its raw structure and its
eligible relaxed endpoint. The raw-group retention is $R_{\rm raw}=K/N$, and for $K>0$ the
retained-group distribution is $q_g=k_g/K$. With training frequencies $p_g$,
the Bhattacharyya coefficient is $\mathrm{BC}(p,q)=\sum_g\sqrt{p_g q_g}$.
Equivalently, in counts,
$\mathrm{SPARC}=100\big(\sum_g\sqrt{p_g k_g/N}\big)^2$, because
\[
\left(\sum_g\sqrt{p_g k_g/N}\right)^2
=\frac{K}{N}\left(\sum_g\sqrt{p_g k_g/K}\right)^2
=R_{\rm raw}\,\mathrm{BC}(p,q)^2.
\]
If no attempt is retained, SPARC is zero.
Because it uses the raw detected group, it also applies to agnostic models.
This retention differs from requested group retention: for Orbitol-X
on MP-20 at 0.1~\AA, the two rates are 76.6\% and 74.7\%,
respectively.

For a fixed group mix, SPARC is linear in $R_{\rm raw}$. Equivalently, it is 100
times the squared affinity between $p$, extended by an empty bin, and the
distribution of outcomes over attempts
$(k_1/N,\ldots,k_{230}/N,1-R_{\rm raw})$, in which every attempt that is not retained
falls into the last bin. By the Cauchy--Schwarz inequality, retained
structures confined to a set $A$ of groups with training share
$p(A)=\sum_{g\in A}p_g$ give
$\mathrm{SPARC}\leq 100\,R_{\rm raw}\,p(A)\leq 100\,p(A)$.
A generator that retains the group of every structure but produces only P1 therefore
scores 1.7 on MP-20. The score reaches 100 only if every attempt retains its
group and the retained-group distribution equals $p$. We use every group
without a minimum count, and all 2,500 attempts remain in the denominator.

At this attempt budget, an ideal sampler that retains the group of every structure and
draws groups from the training distribution has expected scores of 97.9
on MP-20 and 98.3 on Alex-MP-20, estimated as means of 2,000 simulated
sets of 2,500 attempts. These simulation means are reference values,
not hard upper bounds on individual sets of samples. Because the score is concave
in the counts, bootstrap resamples score lower on average than the
original samples. For the three Orbitol-X seeds, the 95\% percentile intervals
average 70.6--74.0 around a mean score of 73.8, so we compare models through
their differences rather than through these intervals.

Replacing the raw detected group by the requested or declared group gives
a variant of SPARC: 72.1 for Orbitol-X, 73.5 for CrystaLLM and 70.6 for
SGFM on MP-20; Orbitol-X and CrystaLLM swap places relative to the raw
variant, with similar observed scores.

\subsection{Additional results and reported comparisons}
\label{apdx:additional_results}

Tables~\ref{tab:full_results} and~\ref{tab:full_results_symmetry} give the
generation and symmetry diagnostics on both datasets, including
agnostic baselines.

\input{tables/v3/full_results}

To estimate relaxation variability, we relaxed the same 500 structures three
times on each compute node, 125 each from
the 400k-step Orbitol-X (MP-20) checkpoint with EDM skip
(Table~\ref{tab:ablations}), SGFM, Crystalite and Orbitol-X (Alex-MP-20). Within a node, up to 17 of 500 structures changed
their detected group at 0.1~\AA, and up to 11 differed in energy by more
than 10~meV/atom; across nodes, the largest counts were 18 and 10. Group changes in the 125-structure subsets reached 5 for
Orbitol-X (MP-20), 8 for SGFM, 6 for Orbitol-X (Alex-MP-20) and zero for
Crystalite, so retention differences of 1--3 percentage points can lie
within this variation. This measures variability for fixed inputs, not over
training or sampling seeds.

\begin{figure}[t]
    \centering
    \includegraphics[width=\linewidth]{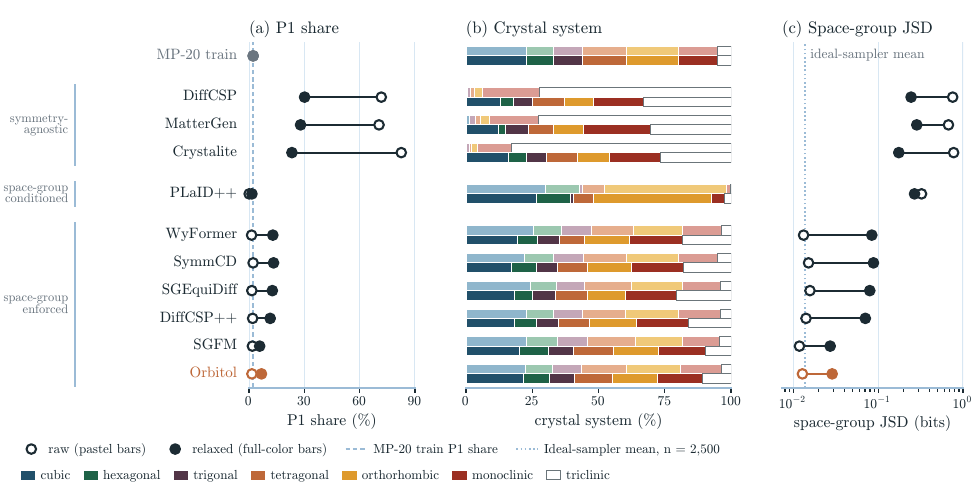}
    \caption{Raw and relaxed symmetry at the stricter positional detection
    tolerance of 0.01~\AA, with angle tolerance $5^\circ$. Models, samples,
    relaxation and endpoint exclusions are the same as in
    Figure~\ref{fig:symmetry_distribution}. Hollow markers and pastel
    bars show raw structures; filled markers and full-color bars show
    relaxed structures. Panels report P1 share on a linear axis,
    crystal-system shares and JSD in bits on a logarithmic axis, using
    MP-20 train detected at the same tolerance. Reference lines are as in
    Figure~\ref{fig:symmetry_distribution}, with the JSD ideal-sampler mean at this
    tolerance (0.0137 bits at 2,500 draws).}
    \label{fig:symmetry_distribution_0p01}
\end{figure}

Figure~\ref{fig:symmetry_distribution_0p01} repeats the main symmetry
comparison at 0.01~\AA. These values depend on detection tolerance as well
as relaxed geometry and do not establish dynamical stability.

Tables~\ref{tab:lemat_metrics} and~\ref{tab:lemat_metrics_alex} give additional
LeMat diagnostics and separate values reported under the authors' own evaluation settings.
The leaderboard includes author-submitted structures, sometimes relaxed
before submission. GEODE reports yields without relaxation and after relaxation
with a different potential, and its MSUN excludes SUN. These reported blocks
therefore cannot be compared directly with ours.

\input{tables/v3/lemat_metrics}

\subsection{Coordinate-configuration comparisons}
\label{apdx:ablations}

\input{tables/ablations}

The main MP-20 model uses the sine skip for its shift-equivariant
parameterization (Appendix~\ref{apdx:implementation_objectives}); the
comparisons in Table~\ref{tab:ablations} do not establish a general yield
advantage. The 400k MP-20 pair has higher observed MSUN with the EDM skip.
The Alex pair changes both input and readout and identifies the Orbitol-I
chart-input application model, but its higher observed MSUN is not a
statistically established benefit of chart input. Neither pair establishes
a yield advantage of orbit representation over Crystalite, of P over M
at matched training.

\FloatBarrier
\subsection{Sampler settings and sensitivity}
\label{apdx:sampler_sweep}

Table~\ref{tab:hyperparameters} gives the samplers of the two benchmark
models. The MP-20 400k controls and the Alex-MP-20 Orbitol-I variant use the
Alex-MP-20 sampler.

Table~\ref{tab:sampler_sweep} varies the sampler of the main MP-20
checkpoint, holding the model fixed. Each setting draws 10,000 attempts
and relaxes every constructed structure with NequIP-OAM-L for 500 FIRE
steps; the in-house evaluator uses an MP2020-like MP-20 reference hull
and novelty against MP-20 train. MSUN is over all attempts, with failures
counting as zero, and is on a different scale from the LeMat results.

The 25-step and 150-step settings have similar observed yield in this
screen. Raising $\sigma_{\min}$ from 0.002 to 0.01 at 25 steps improves
one-shot MSUN and lowers the median initial force, while raising it to
0.05 reduces both one-shot and relaxed MSUN. Ten steps reduce observed
MSUN, and 50 steps offer little change from 25; these observations
support the chosen setting without establishing statistical equivalence
between samplers.

\input{tables/v3/sampler_sweep}

\subsection{Sampling throughput and implementation}
\label{apdx:cost}

\paragraph{Timing setup.}
All times were measured after warm-up on one NVIDIA RTX 6000 Ada (48~GB)
with 8 CPU cores. Baselines used released samplers with the settings and
precision of their quality evaluations in Tables~\ref{tab:mp20_dng}
and~\ref{tab:full_results}. We tuned batch
sizes and removed sample-preserving overheads (batched SGFM, MatterGen
without trajectory recording and CrystalFormer with exact draws per
space group), but did not compile baseline networks or reduce their
step counts. Orbitol uses the sampler settings above, with packing,
compilation and the fast decoder described below.

Timing runs from the prior draw to \lib{pymatgen} structures in memory,
with CUDA synchronized at both ends, per 1,000 attempts including failures.
Model loading, compilation, file writes and relaxation are excluded.
Starting at the batch size of each quality evaluation, we doubled the batch until memory
ran out, a batch exceeded 10 minutes, or two doublings failed to improve
the best time by 5\%; Table~\ref{tab:cost_sampling} reports the fastest
measured batch. For diffusion and flow models, warm network windows
predicted full times and were checked against complete reference batches;
models with errors above 5\% were timed on complete batches at every size,
as was SymmCD because of its non-network overhead. Autoregressive models
used complete generations, and Orbitol used two complete pools of 4,096
attempts per setting because its per-attempt decoding overhead falls with
pool size. Calibration checks before and after each model excluded one
node with CPU contention. Each baseline value is one measurement; each
Orbitol value is the mean of two pools.
The timings combine sampler choices with the implementation below.
They do not isolate the contribution of each component.

\begin{table}[t]
\centering
\small
\caption{Sampling cost on one RTX 6000 Ada, per 1,000 attempts at each model's
batch size with the highest measured throughput. NFE is the
number of network evaluations per structure (token positions for autoregressive models). Speedup is the time
of the model divided by Orbitol's 1.09~s, computed from unrounded times (for ALM Gen, see note b).}
\label{tab:cost_sampling}
\begin{tabular}{@{}lrrrrr@{}}
\toprule
Model & Batch & s per 1,000 & Structures/s & NFE & Speedup \\
\midrule
\multicolumn{6}{@{}l}{\emph{Symmetry agnostic}} \\
Crystalite & 512 & 13.8 & 72 & 299 & 13 \\
DiffCSP & 100 & 260 & 3.8 & 2,000 & 240 \\
MiAD & 64 & 1,218 & 0.82 & 2,000 & 1,123 \\
MatterGen$^{a}$ & 64 & 5,044 & 0.20 & 2,000 & 4,647 \\
ALM Gen$^{b}$ & 16 & 8,377 & 0.12 & 2,000 & 1,482 \\
\addlinespace
\multicolumn{6}{@{}l}{\emph{Space-group conditioned}} \\
CrystaLLM$^{c}$ & 1 & 1,603 & 0.62 & 380 & 1,477 \\
CrystaLLM-$\pi$ & 160 & 104 & 9.6 & 409 & 96 \\
PLaID++$^{d}$ & 1,024 & 156 & 6.4 & 134 & 143 \\
\addlinespace
\multicolumn{6}{@{}l}{\emph{Space-group enforced}} \\
CrystalFormer$^{e}$ & 128 & 215 & 4.7 & 105 & 198 \\
SGEquiDiff & 2,000 & 338 & 3.0 & 1,998 & 312 \\
WyFormer$^{f}$ & 256 & 1,021 & 0.98 & 1,998 & 941 \\
SymmCD$^{k}$ & 1,000 & 195 & 5.1 & 2,000 & 180 \\
DiffCSP++ & 500 & 987 & 1.0 & 1,998 & 909 \\
SGFM, batched$^{g}$ & 512 & 371 & 2.7 & 500 & 341 \\
SGFM, released & 1 & 2,545 & 0.39 & 500 & 2,345 \\
\addlinespace
\multicolumn{6}{@{}l}{\emph{Ours}} \\
Orbitol$^{h}$ & 4,096 & 1.09 & 921 & 49 & 1 \\
Orbitol, reference decoder$^{i}$ & 4,096 & 2.37 & 420 & 49 & 2.2 \\
Orbitol, 150 steps$^{j}$ & 4,096 & 5.40 & 185 & 299 & 5.0 \\
\bottomrule
\end{tabular}
\par\smallskip
\begin{minipage}{\linewidth}
\footnotesize
$^{a}$Trajectory recording turned off.
$^{b}$Alex-MP-20 MatterGen decoder; 2,000 evaluations on a batch doubled by classifier-free guidance. The speedup is
relative to Orbitol's Alex-MP-20 model of width 512 with the 150-step sampler (5.7~s), which we did not re-time
with the 25-step sampler.
$^{c}$The released sampler generates one structure at a time without a key-value cache.
$^{d}$The authors' recipe: 4-bit weights with bfloat16 compute, FlashAttention-2 and batch 1,024,
as in the quality evaluation, which decoded serially.
$^{e}$Float64 as released; exact draws per space group of the sampling plan. JAX compiles once per space
group and batch shape (about 17~s each), which is excluded.
$^{f}$Wyckoff genes sampled on the CPU, then DiffCSP++ geometry; both stages are timed.
$^{g}$Our batched reimplementation of the released sampling loop, checked against it.
$^{h}$The Orbitol-X model of Table~\ref{tab:mp20_dng}, with the 25-step sampler of
Appendix~\ref{apdx:sampler_sweep}; packed, compiled, bfloat16, fast decoder; mean of two complete pools.
$^{i}$The reference decoder, which decoded the scored pools; both decoders produce the same structures (fractional
coordinates agree within $10^{-7}$).
$^{j}$The 150-step sampler of Appendix~\ref{apdx:sampler_sweep} on the 400k-step checkpoint
of the same architecture.
$^{k}$Complete batches; the window prediction at the same batch agrees within 0.2\%.
\end{minipage}
\end{table}

\phantomsection\label{apdx:speed}
\paragraph{Short sequences and packing.}
Orbitol uses one token per orbit and one lattice token. MP-20 structures
have 4.70 orbits on average against 10.4 atoms in their stored cells,
roughly halving the token count and quartering the attention pairs.
Packing avoids padding every crystal to 21 positions.
The sampler groups whole crystals in request order into fixed chunks of
$T=8{,}192$ tokens at width 512, with about 1,430 MP-20 crystals per chunk.
FlexAttention's document mask isolates crystals, and GEM bias is indexed
by crystal, head and token pair. Layouts and masks are reused across all
network evaluations; unused tokens attend only to themselves, while
integration states and noise draws retain padded shapes
(Appendix~\ref{apdx:implementation_sampling}). Packing needs large pools:
with 256 attempts, the 25-step sampler takes 3.4~s per 1,000.

\paragraph{Compilation.}
At width 512, \texttt{torch.compile} in \texttt{reduce-overhead} mode
captures the fixed-shape transformer trunk as a CUDA graph reused across
chunks. Embeddings, GEM geometry and output heads remain outside the
compiled trunk. The network uses bfloat16, while integration states,
chart arithmetic and GEM geometry use single precision. Compilation and
graph capture take about 17~s per process and are excluded from the timings.

\paragraph{Decoding.}
The fast decoder caches Wyckoff tables, expands orbit members in one array
operation, and compiles collision and symmetry-containment checks with
\lib{Numba}. Niggli and LLL reduction follow \lib{pymatgen}'s arithmetic,
with fallback to the reference decoder for uncertified reductions or
structures near numerical boundaries. On 30,000 stored outputs and four
live runs of 10,000 attempts, atom types, masks and lattices agreed
bitwise with the reference decoder used for the scored pools, and
fractional coordinates agreed within $10^{-7}$. The fast path benefits
batches; its fixed cost makes it slower for a single structure.

\paragraph{Step count and numerical equivalence.}
The 5.40 to 1.09~s comparison changes the checkpoint and noise settings
as well as step count, so it does not isolate a step-count speedup;
the 150-step configuration is still about 60 times faster than SGEquiDiff.
Complete-pool timing includes prior draws, decoding and structure
construction, whose per-attempt overhead depends on pool size.
Packing and compilation compute the same quantities with different
kernels and operation order, so floating-point differences can grow
through bfloat16 sampling trajectories until individual decoded structures
differ. Decoder agreement on fixed sampled states therefore does not
imply bitwise agreement between sampling trajectories. The MP-20 pools of
Table~\ref{tab:mp20_dng} were sampled with the same packed, compiled sampler;
only their decoder differs.

%% file: tables/v3/full_results.tex
\begin{table*}[t]
\centering
\small
\caption{Generation and discovery diagnostics for all models on MP-20 and Alex-MP-20. Values are from 2,500 sampled attempts per model; the MP-20 Orbitol-X and PLaID++ rows are means over three sampling seeds, each metric averaged per seed. Valid, Unique and Novel are separate LeMat percentages of attempts; Unique counts valid structures unique among the model's attempts, and Novel counts valid structures absent from LeMat-Bulk. SUN and MSUN are joint unique and novel yields with $E_{\mathrm{hull}}\le 0$ and $E_{\mathrm{hull}}\le 0.1$~eV/atom, respectively, as percentages of attempts. Raw P1 is the percentage of detected structures in P1 before relaxation at $\texttt{symprec}=0.1$~\AA. Training P1: MP-20 $1.7\%$, Alex-MP-20 $0.7\%$. Bold marks column maxima within each dataset, except P1. Agnostic models have no LeMat scores in our evaluation. Metric details are in Section~\ref{sec:exp_protocol}. Symmetry after relaxation is in Table~\ref{tab:full_results_symmetry}.}
\label{tab:full_results}
\setlength{\tabcolsep}{5pt}
\begin{tabular}{@{}l rrr rr r@{}}
\toprule
& \multicolumn{3}{c}{Generation (\%)} & \multicolumn{2}{c}{Discovery (\%)} & \\
\cmidrule(lr){2-4}\cmidrule(lr){5-6}
Model & Valid & Unique & Novel & SUN & MSUN & Raw P1 (\%) \\
\midrule
\multicolumn{7}{@{}l}{\textbf{MP-20}} \\
\multicolumn{7}{@{}l}{\emph{Symmetry agnostic}} \\
Crystalite$^{a}$ & \NA & \NA & \NA & \NA & \NA & 30.3 \\
DiffCSP$^{b}$ & \NA & \NA & \NA & \NA & \NA & 25.8 \\
MatterGen$^{b}$ & \NA & \NA & \NA & \NA & \NA & 20.9 \\
MiAD$^{b}$ & \NA & \NA & \NA & \NA & \NA & 53.7 \\
\addlinespace
\multicolumn{7}{@{}l}{\emph{Space group conditioned}} \\
CrystaLLM & 95.6 & 87.2 & 21.3 & 1.00 & 10.4 & 2.4 \\
CrystaLLM-$\pi$ & 95.0 & 92.4 & 31.9 & \best{1.40} & 16.1 & 1.6 \\
PLaID++ & 92.1 & 74.3 & 25.0 & 0.95 & 12.0 & 0.1 \\
PLaID++, released outputs$^{\dagger}$ & 98.8 & 78.7 & 28.1 & 1.16 & 13.3 & 0.0 \\
\addlinespace
\multicolumn{7}{@{}l}{\emph{Space group enforced}} \\
CrystalFormer & 93.6 & 92.5 & 51.0 & 0.80 & 13.9 & 2.2 \\
SGEquiDiff & 96.0 & 95.0 & 57.1 & 0.68 & 18.1 & 1.5 \\
WyFormer & 91.4 & 90.9 & 59.1 & 0.40 & 14.2 & 1.4 \\
SymmCD & 92.9 & 92.0 & 58.8 & 0.84 & 15.9 & 2.2 \\
DiffCSP++ & 95.7 & 95.1 & \best{63.6} & 0.60 & 16.2 & 2.0 \\
SGFM & \best{97.1} & \best{95.9} & 42.1 & 1.00 & 16.9 & 1.8 \\
\addlinespace
\multicolumn{7}{@{}l}{\emph{Ours (space group enforced)}} \\
Orbitol-X & 96.5 & 95.4 & 47.3 & 1.11 & \best{19.0} & 1.6 \\
\midrule
\multicolumn{7}{@{}l}{\textbf{Alex-MP-20}} \\
\multicolumn{7}{@{}l}{\emph{Symmetry agnostic}} \\
ALM Gen$^{c}$ & \NA & \NA & \NA & \NA & \NA & 8.4 \\
\addlinespace
\multicolumn{7}{@{}l}{\emph{Ours (space group enforced)}} \\
Orbitol-X, $d=512$, 650k & 93.9 & 93.9 & 60.7 & 0.68 & 34.2 & 0.6 \\
\bottomrule
\end{tabular}
\par\smallskip
\begin{minipage}{\textwidth}
\footnotesize
Checkpoint, retraining and PLaID++ sampling details are in Table~\ref{tab:mp20_dng}.
$^{a}$Our reproduction of the released Crystalite recipe (1.07M steps).
$^{b}$Our draw from the released checkpoint.
$^{c}$Alex-MP-20 MatterGen decoder, released placeholder prompt, guidance 0.5.
$^{\dagger}$The authors' released pool of the same model, first 2,500 rows. Their sampler discards generations that fail to decode and samples again, so its rates are over decoded rows, not attempts; shown for reference and not bolded.
\end{minipage}
\end{table*}

\begin{table*}[t]
\centering
\small
\caption{Symmetry after relaxation for all models on MP-20 and Alex-MP-20. Values are from 2,500 sampled attempts per model; the MP-20 Orbitol-X and PLaID++ rows are means over three sampling seeds, each metric averaged per seed. Retention is the percentage of attempts whose relaxed group equals the requested or declared group. SPARC combines retention of the raw detected group with coverage of the training distribution, on a scale from 0 to 100 (Equation~\ref{eq:sparc}). P1 is the percentage of detected relaxed structures in P1; JSD compares their group histogram with the training histogram of the same dataset, in bits. Detection uses $\texttt{symprec}=0.1$~\AA; failed relaxations and implausible endpoints are excluded from the histograms and count as losses for retention and SPARC. Training P1: MP-20 $1.7\%$, Alex-MP-20 $0.7\%$. Bold marks maximum retention and SPARC and minimum JSD within each dataset. Agnostic models have no requested group. Metric details are in Section~\ref{sec:exp_protocol}.}
\label{tab:full_results_symmetry}
\setlength{\tabcolsep}{5pt}
\begin{tabular}{@{}l rrrr@{}}
\toprule
& \multicolumn{4}{c}{After relaxation} \\
\cmidrule(l){2-5}
Model & \shortstack{Requested group\\retained (\%)} & SPARC & P1 (\%) & JSD (bits) \\
\midrule
\multicolumn{5}{@{}l}{\textbf{MP-20}} \\
\multicolumn{5}{@{}l}{\emph{Symmetry agnostic}} \\
Crystalite$^{a}$ & \NA & 39.0 & 21.5 & 0.177 \\
DiffCSP$^{b}$ & \NA & 34.6 & 28.3 & 0.247 \\
MatterGen$^{b}$ & \NA & 40.2 & 26.4 & 0.290 \\
MiAD$^{b}$ & \NA & 1.7 & 8.9 & 0.099 \\
\addlinespace
\multicolumn{5}{@{}l}{\emph{Space group conditioned}} \\
CrystaLLM & \best{77.8}$^{d}$ & 73.5 & 4.7 & 0.047 \\
CrystaLLM-$\pi$ & 74.3$^{d}$ & 69.4 & 4.6 & 0.038 \\
PLaID++ & 76.6$^{d}$ & 36.3 & 1.3 & 0.267 \\
PLaID++, released outputs$^{\dagger}$ & 81.2$^{d}$ & 39.7 & 1.6 & 0.257 \\
\addlinespace
\multicolumn{5}{@{}l}{\emph{Space group enforced}} \\
CrystalFormer & 62.3 & 59.4 & 11.1 & 0.071 \\
SGEquiDiff & 54.7 & 51.1 & 11.0 & 0.077 \\
WyFormer & 51.8 & 48.6 & 11.2 & 0.081 \\
SymmCD & 57.4 & 54.5 & 11.1 & 0.085 \\
DiffCSP++ & 57.3 & 55.1 & 9.7 & 0.069 \\
SGFM & 73.4 & 71.9 & 4.9 & \best{0.027} \\
\addlinespace
\multicolumn{5}{@{}l}{\emph{Ours (space group enforced)}} \\
Orbitol-X & 74.7 & \best{73.8} & 5.6 & 0.029 \\
\midrule
\multicolumn{5}{@{}l}{\textbf{Alex-MP-20}} \\
\multicolumn{5}{@{}l}{\emph{Symmetry agnostic}} \\
ALM Gen$^{c}$ & \NA & 62.8 & 12.3 & 0.155 \\
\addlinespace
\multicolumn{5}{@{}l}{\emph{Ours (space group enforced)}} \\
Orbitol-X, $d=512$, 650k & 71.6 & \best{71.0} & 4.4 & \best{0.043} \\
\bottomrule
\end{tabular}
\par\smallskip
\begin{minipage}{\textwidth}
\footnotesize
Checkpoint, retraining and PLaID++ sampling details are in Table~\ref{tab:mp20_dng}.
$^{a}$Our reproduction of the released Crystalite recipe (1.07M steps).
$^{b}$Our draw from the released checkpoint.
$^{c}$Alex-MP-20 MatterGen decoder, released placeholder prompt, guidance 0.5.
$^{\dagger}$The authors' released pool of the same model, first 2,500 rows. Their sampler discards generations that fail to decode and samples again, so its rates are over decoded rows, not attempts; shown for reference and not bolded.
An ideal sampler of the training distribution with 2,500 draws has a JSD of 0.0137~bits on average (5th--95th percentile range 0.0114--0.0160; Figure~\ref{fig:symmetry_distribution}). The Orbitol--SGFM JSD difference is small relative to this finite-sample spread; additional samples would be needed to resolve it.
$^{d}$Uses the group declared by the model.
\end{minipage}
\end{table*}

%% file: tables/v3/lemat_metrics.tex
\begin{table*}[t]
\centering
\small
\caption{LeMat-GenBench metrics on MP-20. Our evaluation is that of Table~\ref{tab:mp20_dng}: 2,500 attempts per model, NequIP-OAM-L pre-relaxation, LeMat-GenBench with raw fallback. The leaderboard blocks are its MP-20 view, snapshot of 2026-10-06 \citep{betala2026lematgenbenchunifiedevaluationframework}, grouped by the submission's pre-relaxation flag; the leaderboard scores the structures that authors submit. The GEODE block is its Table~1 \citep{Lou2026GEODE}. Scores are comparable only within a block. Values are as served or printed, except MSUN in our and the leaderboard blocks, which includes SUN (for the leaderboard, SUN plus its metastable column). Meta.\ denotes metastable. Bold marks the best value in each block (lowest $E_{\mathrm{hull}}$).}
\label{tab:lemat_metrics}
\setlength{\tabcolsep}{5pt}
\begin{tabular}{@{}l rrrrrrr r@{}}
\toprule
Model & Valid & Unique & Novel & Stable & Meta. & SUN & MSUN & $E_{\mathrm{hull}}$ \\
 & (\%) & (\%) & (\%) & (\%) & (\%) & (\%) & (\%) & (eV/atom) \\
\midrule
\multicolumn{9}{@{}l}{\emph{Our evaluation (Table~\ref{tab:mp20_dng})}} \\
CrystaLLM & 95.6 & 87.2 & 21.3 & \best{11.3} & \best{71.1} & 1.0 & 10.4 & \best{0.052} \\
CrystaLLM-$\pi$ & 95.0 & 92.4 & 31.9 & 8.6 & 67.1 & \best{1.4} & 16.1 & 0.066 \\
PLaID++ & 92.1 & 74.3 & 25.0 & 7.8 & 67.1 & 0.9 & 12.0 & 0.061 \\
CrystalFormer & 93.6 & 92.5 & 51.0 & 5.8 & 46.6 & 0.8 & 13.9 & 0.153 \\
SGEquiDiff & 96.0 & 95.0 & 57.1 & 4.7 & 47.2 & 0.7 & 18.1 & 0.137 \\
WyFormer & 91.4 & 90.9 & 59.1 & 3.5 & 37.3 & 0.4 & 14.2 & 0.161 \\
SymmCD & 92.9 & 92.0 & 58.8 & 3.7 & 40.6 & 0.8 & 15.9 & 0.173 \\
DiffCSP++ & 95.7 & 95.1 & \best{63.6} & 3.6 & 38.4 & 0.6 & 16.2 & 0.176 \\
SGFM & \best{97.1} & \best{95.9} & 42.1 & 7.4 & 58.3 & 1.0 & 16.9 & 0.097 \\
Orbitol & 96.5 & 95.4 & 47.3 & 6.7 & 56.4 & 1.1 & \best{19.0} & 0.108 \\
\addlinespace
\multicolumn{9}{@{}l}{\emph{Leaderboard, pre-relaxed submissions}} \\
Crystalite & \best{97.2} & 95.8 & 53.2 & 12.7 & 51.6 & 1.5 & \best{24.1} & 0.090 \\
MCFlow & \best{97.2} & \best{96.3} & 52.2 & 11.9 & 49.3 & 0.7 & 19.6 & 0.099 \\
OMatG & 96.4 & 95.2 & 51.2 & 11.6 & 49.8 & 1.0 & 19.0 & 0.096 \\
MiAD & 96.2 & 94.3 & 40.2 & 6.4 & \best{62.0} & 1.0 & 17.6 & 0.080 \\
MatterGen & 95.7 & 95.1 & \best{70.5} & 2.0 & 33.4 & 0.2 & 15.2 & 0.183 \\
OMatG-FC & \best{97.2} & 92.8 & 28.9 & \best{18.4} & 57.9 & \best{1.7} & 13.7 & \best{0.069} \\
Zatom-1-WD & 88.3 & 86.0 & 37.9 & 4.7 & 50.5 & 0.3 & 9.1 & 0.109 \\
PLaID++ & 96.0 & 77.8 & 24.2 & 12.4 & 60.7 & 1.0 & 8.6 & 0.085 \\
WyFormer-DFT & 95.2 & 95.0 & 66.4 & 3.7 & 24.8 & 0.4 & 8.2 & 0.271 \\
WyFormer & 93.4 & 93.0 & 66.4 & 0.5 & 15.7 & 0.1 & 2.0 & 0.499 \\
\addlinespace
\multicolumn{9}{@{}l}{\emph{Leaderboard, submissions without pre-relaxation}} \\
Chemeleon2 & 95.2 & 88.1 & \best{71.6} & 0.0 & 39.8 & 0.0 & \best{21.2} & \best{0.156} \\
Chemeleon1 & 95.4 & 94.8 & 64.0 & \best{3.6} & 34.4 & \best{0.3} & 11.6 & 0.205 \\
DiffCSP & \best{95.7} & 94.8 & 66.2 & 2.3 & 29.8 & 0.1 & 8.6 & 0.275 \\
DiffCSP++ & 95.3 & \best{95.1} & 62.0 & 1.0 & 26.4 & 0.2 & 5.2 & 0.409 \\
CrystaLLM-$\pi$ & 86.8 & 84.9 & 24.9 & 2.9 & \best{49.0} & \best{0.3} & 3.9 & 0.321 \\
SymmCD & 73.4 & 73.0 & 47.0 & 1.4 & 18.6 & 0.1 & 2.5 & 0.876 \\
LLaMat2 & 84.4 & 81.4 & 30.0 & 0.7 & 34.7 & 0.1 & 2.2 & 0.440 \\
CrystalFormer & 69.9 & 69.4 & 31.8 & 1.4 & 28.8 & 0.0 & 3.1 & 0.704 \\
ADiT & 90.6 & 87.8 & 26.0 & 0.4 & 36.5 & 0.0 & 1.0 & 0.333 \\
Zatom-1-WD & 67.2 & 65.4 & 24.2 & 0.2 & 22.8 & 0.0 & 0.4 & 0.519 \\
LLaMat3 & 15.4 & 15.2 & 10.5 & 0.1 & 2.1 & 0.0 & 0.2 & 1.707 \\
Crystal-GFN & 51.7 & 51.7 & 51.7 & 0.0 & 0.0 & 0.0 & 0.0 & 2.086 \\
\addlinespace
\multicolumn{9}{@{}l}{\emph{Reported by GEODE$^{a}$}} \\
GEODE & 95.8 & 94.9 & 55.8 & \NA & 41.9 & \NA & 11.9 & 0.196 \\
GEODE$^{*}$ & 95.6 & 94.9 & 54.3 & \NA & \best{43.7} & \NA & 16.4 & \best{0.135} \\
GEODE, filtered templates$^{*}$ & \best{96.4} & \best{95.3} & \best{74.0} & \NA & 38.8 & \NA & \best{23.8} & 0.162 \\
\bottomrule
\end{tabular}
\par\smallskip
\begin{minipage}{\textwidth}
\footnotesize
For the six symmetry-aware models in our evaluation and on the leaderboard, our evaluation gives measured MSUN rates 3.4--13.4 percentage points higher. Some leaderboard scores in the snapshot were computed before the benchmark's MACE checkpoint fix. $^{a}$GEODE Table~1 (arXiv v1); $^{*}$structures relaxed with EquiformerV3 before scoring. Its MSUN column excludes SUN, which it does not print; GEODE states that every row has a SUN rate of at most 2.12\%.
\end{minipage}
\end{table*}

\begin{table*}[t]
\centering
\small
\caption{LeMat-GenBench metrics on Alex-MP-20, as in Table~\ref{tab:lemat_metrics}. The GEODE blocks are its Table~1 \citep{Lou2026GEODE}; its Alex-MP-20 baselines are copies of the LeMat-GenBench leaderboard. LeMat-Bulk contains Alexandria, so novelty is measured against a reference that overlaps the training source. Scores are comparable only within a block.}
\label{tab:lemat_metrics_alex}
\setlength{\tabcolsep}{5pt}
\begin{tabular}{@{}l rrrrrrr r@{}}
\toprule
Model & Valid & Unique & Novel & Stable & Meta. & SUN & MSUN & $E_{\mathrm{hull}}$ \\
 & (\%) & (\%) & (\%) & (\%) & (\%) & (\%) & (\%) & (eV/atom) \\
\midrule
\multicolumn{9}{@{}l}{\emph{Our evaluation (Table~\ref{tab:full_results})}} \\
Orbitol & 93.9 & 93.9 & 60.7 & 2.0 & 59.4 & 0.7 & 34.2 & 0.101 \\
\addlinespace
\multicolumn{9}{@{}l}{\emph{Reported by GEODE$^{a}$}} \\
GEODE & \best{95.0} & \best{94.9} & 54.1 & \NA & 57.7 & \NA & 25.8 & 0.102 \\
GEODE$^{*}$ & \best{95.0} & \best{94.9} & 52.5 & \NA & \best{61.8} & \NA & 29.2 & \best{0.084} \\
GEODE, filtered templates$^{*}$ & 94.8 & 94.6 & \best{64.2} & \NA & 61.7 & \NA & \best{36.8} & 0.086 \\
\addlinespace
\multicolumn{9}{@{}l}{\emph{Leaderboard as copied by GEODE$^{a}$, without pre-relaxation}} \\
Chemeleon 2 & \best{97.4} & 93.5 & \best{93.7} & \NA & 67.4 & \NA & \best{62.4} & 0.096 \\
Chemeleon 1 & 94.0 & \best{93.9} & 59.8 & \NA & 54.4 & \NA & 28.0 & 0.112 \\
CrystaLLM-$\pi$ & 93.4 & 93.2 & 23.9 & \NA & \best{71.3} & \NA & 12.5 & \best{0.094} \\
\bottomrule
\end{tabular}
\par\smallskip
\begin{minipage}{\textwidth}
\footnotesize
$^{a}$GEODE Table~1 (arXiv v1); $^{*}$structures relaxed with EquiformerV3 before scoring. Its MSUN column excludes SUN, which it does not print; GEODE states that every row has a SUN rate of at most 2.12\%.
\end{minipage}
\end{table*}

%% file: tables/ablations.tex
\begin{table*}[!htbp]
\centering
\small
\caption{Coordinate configurations under our common evaluation:
2,500 attempts, 500 unconstrained NequIP-OAM-L FIRE steps, and LeMat-GenBench
scoring with raw fallback (Appendix~\ref{apdx:evaluation}). The MP-20 pair shares target P and changes
the coordinate skip; all four rows use distance GEM. The datasets form
separate comparisons. SUN and MSUN are percentages of all attempts, and
MSUN includes SUN. Each result is one trained model and one sampling seed;
matching training updates does not control random variation.}
\label{tab:ablations}
\setlength{\tabcolsep}{6pt}
\begin{tabular}{@{}llrrr@{}}
\toprule
Model & Coordinate configuration & Training updates & SUN (\%) & MSUN (\%) \\
\midrule
\multicolumn{5}{@{}l}{\emph{MP-20, coordinate skip; type weight 1}} \\
Orbitol-X & Representative / projection, P, sine skip & 400k & 1.24 & 17.1 \\
Orbitol-X & Representative / projection, P, EDM skip & 400k & 1.16 & 18.9 \\
\addlinespace
\multicolumn{5}{@{}l}{\emph{Alex-MP-20, input and readout; type weight 0.5}} \\
Orbitol-X & Representative / projection, P, sine skip & 650k & 0.68 & 34.2 \\
Orbitol-I & Chart / direct, P, sine skip & 650k & 0.72 & 36.6 \\
\bottomrule
\end{tabular}
\par\smallskip
\begin{minipage}{\textwidth}
\footnotesize
All models have width 512 and use EMA weights. MP-20 checkpoints are from
500k-update schedules; Alex checkpoints are from 1M-update schedules.
All four use the 150-step sampler (Table~\ref{tab:experiment_configs} and
Appendix~\ref{apdx:sampler_sweep}). The MP-20 pair differs from the main
1.5M-update model, whose type weight is 0.25.
\end{minipage}
\end{table*}

%% file: tables/v3/sampler_sweep.tex
\begin{table}[t]
\centering
\small
\setlength{\tabcolsep}{4pt}
\caption{Sampler sweep on the MP-20 Orbitol-X model of Table~\ref{tab:mp20_dng} (in-house evaluation against the MP-20 hull and training set; 10,000 attempts per setting; every constructed structure relaxed with NequIP-OAM-L for 500 FIRE steps). MSUN scores the relaxed structures and one-shot MSUN the raw structures at their unrelaxed NequIP energy; both are over all attempts, with failures counting as zero. UN is the unique and novel fraction of the raw structures, over constructed structures. $F_0$ is the median initial maximum force (eV/\AA) over successful relaxations. $\gamma$ is the churn per step, $\min(S_{\mathrm{churn}}/N, \sqrt{2}-1)$; churn acts for noise levels above $S_{\min}$; Heun with $N$ steps uses $2N-1$ network evaluations. $S_{\mathrm{noise}}=1.003$ throughout.}
\label{tab:sampler_sweep}
\begin{tabular}{@{}l rrrrrr rrrr@{}}
\toprule
& $\sigma_{\max}$ & $\sigma_{\min}$ & $N$ & $\gamma$ & $S_{\min}$ & $\rho$ & MSUN (\%) & UN (\%) & One-shot (\%) & $F_0$ \\
\midrule
\multicolumn{11}{@{}l}{\emph{Noise range and steps}} \\
checkpoint$^{a}$ & 80 & 0.002 & 150 & 0.4 & 0 & 7 & 44.3 & 74.5 & 30.8 & 0.52 \\
 & 3 & 0.002 & 150 & 0.4 & 0 & 7 & 43.7 & 73.2 & 30.5 & 0.51 \\
 & 3 & 0.002 & 25 & 0.4 & 0 & 7 & 44.3 & 77.1 & 30.2 & 0.54 \\
\addlinespace
\multicolumn{11}{@{}l}{\emph{Endpoint noise at 25 steps}} \\
chosen$^{a}$ & 3 & 0.01 & 25 & 0.4 & 0 & 7 & 44.7 & 76.5 & 31.6 & 0.44 \\
 & 3 & 0.05 & 25 & 0.4 & 0 & 7 & 43.4 & 77.2 & 26.9 & 0.69 \\
\addlinespace
\multicolumn{11}{@{}l}{\emph{Steps at the chosen noise range}} \\
 & 3 & 0.01 & 10 & 0.4 & 0 & 7 & 43.1 & 78.5 & 29.9 & 0.48 \\
 & 3 & 0.01 & 15 & 0.4 & 0 & 7 & 44.2 & 78.2 & 30.9 & 0.46 \\
 & 3 & 0.01 & 50 & 0.4 & 0 & 7 & 44.9 & 75.6 & 31.9 & 0.43 \\
\bottomrule
\end{tabular}
\par\smallskip
\begin{minipage}{\linewidth}\footnotesize
$^{a}$Mean of three sampling seeds; all other rows use one seed. The checkpoint row uses its stored sampler; the chosen row is used for the MP-20 Orbitol-X rows of the main tables.
\end{minipage}
\end{table}

%% file: sections/appendix_piezo.tex
\section{Piezoelectric screening and DFT validation}
\label{apdx:piezo}

\paragraph{Selection.}
The 62 structures were generated by Orbitol models trained on the
Alex-MP-20 and MP-20 datasets and selected for a high predicted figure of
merit. Four of them had PBE band gaps of at most 0.03~eV and were excluded
from the response calculations.

\paragraph{Settings.}
All calculations used \lib{VASP} 6.4.2 \citep{Kresse1996VASP} with the PBE
functional \citep{Perdew1996PBE}, projector augmented-wave potentials (the
PBE\_54 set) \citep{Blochl1994PAW,Kresse1999PAW}, a plane-wave cutoff of
520~eV, aspherical gradient corrections inside the augmentation spheres and
Gaussian smearing of 0.05~eV. We applied the Hubbard $U$ values of the
Materials Project wherever it uses them (12 materials). The eight Mn-, Fe-
and Co-containing materials were computed spin-polarized with ferromagnetic
initial moments, as were the band-gap checks of the four structures
excluded as metals; other magnetic orders were not considered.
$\Gamma$-centred or Monkhorst--Pack $k$-point meshes with a density of 100
points per \AA$^{-3}$ of reciprocal volume were generated with \lib{pymatgen}
\citep{Ong2013Pymatgen}, and each material used the same mesh in all
calculations. Cell shape, volume and atomic positions were relaxed until
all forces fell below 0.005~eV/\AA, with an electronic convergence
criterion of $10^{-6}$~eV ($10^{-7}$~eV for seven materials). The
relaxations started from the symmetrized generated structures without
perturbation and kept their symmetry, which \lib{VASP} imposes by default
(positional tolerance $10^{-5}$~\AA), so they cannot lower the space group.
Symmetry-lowering distortions are tested instead by the $\Gamma$-point
phonons of the DFPT runs, which contain modes of every symmetry. For 15 of
the 58 candidates we also repeated the relaxation without symmetry from
atoms displaced randomly by 0.02~\AA; all 15 returned to their space
group, with energy differences of at most 0.04~meV/atom. One candidate had a
single unstable $\Gamma$ mode in its DFT-relaxed $Cmc2_1$ structure
($40i$~cm$^{-1}$), with atomic displacements perpendicular to the polar axis
that remove the mirror planes. We displaced the atoms along this mode
(largest displacement 0.05~\AA) in the same 18-atom cell and relaxed cell
and atoms again with the same settings. The structure relaxed into $P2_1$,
0.42~meV/atom below $Cmc2_1$ (largest atomic shift 0.18~\AA), with no
unstable $\Gamma$ mode and a positive-definite but soft elastic tensor
(smallest eigenvalue 0.72~GPa). The dielectric
and piezoelectric tensors and the Born effective charges were computed with
DFPT \citep{Wu2005DFPT}, except for two candidates, for
which we used finite differences with atomic displacements of 0.015~\AA.
The elastic tensor was obtained from the stress response to finite lattice
distortions, including the relaxation of the ions, again with atomic
displacements of 0.015~\AA\ and an electronic convergence criterion of
$10^{-7}$~eV. All tensors refer to the same relaxed structure. We detected
the space group of the relaxed structures with \lib{spglib} at a positional
tolerance of $10^{-3}$~\AA, which gave the same groups as $10^{-2}$~\AA.

\paragraph{Stability.}
A candidate is stable at $\Gamma$ if the $\Gamma$-point phonons of its DFPT
run have no imaginary optical mode. One candidate fails with
$18.8i$~cm$^{-1}$, and another with a polar mode at zero frequency within
numerical accuracy ($6.5i$~cm$^{-1}$ in DFPT, $9i$~cm$^{-1}$ from finite
differences), which makes its response values unreliable. It is mechanically stable if its
elastic tensor is positive definite; two candidates fail, one of which also
fails at $\Gamma$, so their piezoelectric strain coefficients are undefined. This test at
$\Gamma$ does not establish dynamical stability over the Brillouin zone. The
two largest figures of merit, 308 and 109 (in $10^{-12}$~m$^2$\,N$^{-1}$), belong to structures near an
elastic instability, with
smallest eigenvalues of the elastic tensor of 0.72 and 1.25~GPa; a nearly
singular elastic tensor inflates $d=e\,C^{-1}$.

\paragraph{Figure of merit.}
We computed $d=e\,C^{-1}$ from the piezoelectric stress tensor $e$ and the
relaxed-ion elastic tensor $C$, with engineering shear strains. The figure
of merit is
\begin{equation}
d_{31}g_{31} = \max_{p,\;u\perp p}
\frac{d_{puu}^{2}}{\varepsilon_0\,p^{\top}\varepsilon^{T}p},
\label{eq:d31g31}
\end{equation}
over unit vectors $p$ (the polarization and electrode normal) and $u$ (the
in-plane strain axis), where $\varepsilon^{T}=\varepsilon^{S}+d\,e^{\top}/\varepsilon_0$
is the permittivity at fixed stress, as in the conventional voltage
coefficient $g$, and $\varepsilon^{S}$ is the static DFPT permittivity at
fixed strain. The figure of merit is the transverse open-circuit response
of a thin plate with electrodes normal to $p$ under uniaxial stress along
$u\perp p$, all other stress components being zero: the electrodes allow
only a field along $p$, and $D\cdot p=0$ then gives
$g_{puu}=d_{puu}/(\varepsilon_0\,p^{\top}\varepsilon^{T}p)$, so
$d_{puu}g_{puu}=d_{puu}^{2}/(\varepsilon_0\,p^{\top}\varepsilon^{T}p)$, which
we maximize over $p$ and $u$. Fringing fields are neglected. The maximum
removes the dependence on the standard IEEE
orientation, which places $z$ along the shortest lattice vector of an
orthorhombic cell rather than along its polar axis and can give $d_{31}=0$
for polar $mm2$ materials. With $\varepsilon^{S}$ in place of $\varepsilon^{T}$,
all values are higher and the same 13 candidates exceed the corresponding
MP 90th percentile (3.70), but the implied coupling factor of the cut,
$k^2=d_{31}g_{31}/s_{uuuu}$ with the compliance $s=C^{-1}$, exceeds one for
these two candidates (8.2 and 1.3), which is impossible for a stable
crystal; with $\varepsilon^{T}$ it stays below one for all 56 mechanically
stable candidates (at most 0.88). DFPT gives
intrinsic single-crystal responses at 0~K, without the extrinsic
contributions of poled ceramics, so we compare only with other DFT values.

\paragraph{Materials Project reference.}
We used the MP release of 2026-09-28: 704 materials with piezoelectric,
dielectric and elastic tensors
\citep{deJong2015Piezo,Petousis2017Dielectric,deJong2015Elastic}, all with a
positive-definite elastic tensor, and evaluated Equation~\ref{eq:d31g31} on
their tensors. We kept the 702 with a non-zero piezoelectric tensor
($\max_{ij}|e_{ij}|\geq10^{-10}$~C\,m$^{-2}$); the two that drop out belong
to class 432. MP's tensors come from its own workflows, so the comparison is
between two DFT data sets with similar but not identical settings. With
all 704 materials the 90th percentile is 3.08 and the count above it is
unchanged.

\paragraph{Energy above the hull.}
For each candidate we recomputed every competing MP and Alexandria PBE
phase with the settings above and applied the MP2020 corrections
\citep{Wang2021MP2020}. The hull excludes Alexandria phases with O--O or
N--N distances below 1.6~\AA, which the per-anion correction makes too
stable; including them lowers the hull and raises five hull energies by
2--17~meV/atom. Hull energies of the ferromagnetic Mn, Fe and Co compounds
are upper bounds. We did not compute hull energies for the two mechanically
unstable candidates.

\paragraph{Novelty.}
We compared each candidate with every MP and Alexandria PBE entry of the
same reduced formula using \lib{pymatgen}'s \texttt{StructureMatcher}, at
its default tolerances (fractional length 0.2, site 0.3, angle $5^\circ$,
volume scaled). We count a match as the same structure only if the space
groups agree and a strict match (length 0.05, site 0.1, angle $2^\circ$)
holds. Of the 55, six have an identical structure in Alexandria. Three match an entry only in a different space group ($R3$ against $Fm\bar{3}m$ and $C2/m$, polar $Fmm2$ against
centrosymmetric $I4/mmm$, and $P3$ against $P1$). The other 46 match no entry even at the default matcher tolerances.